\documentclass[11pt]{amsart}
\usepackage{amsmath, amssymb, amsthm, amsfonts, enumerate, verbatim, mathtools,bbm,esint,bm}
\usepackage{multirow}
\usepackage{enumitem}
\usepackage[normalem]{ulem}
\usepackage{pgf,tikz,pgfplots}
\usepackage{tikz-3dplot}
\usetikzlibrary{shapes,decorations,arrows,calc,arrows.meta,fit,positioning,intersections}

\usepackage{physics}
\usepackage{ifthen}
\usepackage[outline]{contour} 
\usetikzlibrary{angles,quotes} 
\usetikzlibrary{patterns}
\tikzset{>=latex} 
\contourlength{1.2pt}
\colorlet{myred}{red!65!black}
\tikzstyle{ground}=[preaction={fill,top color=black!10,bottom color=black!5,shading angle=20},
fill,pattern=north east lines,draw=none,minimum width=0.3,minimum height=0.6]
\tikzstyle{mass}=[line width=0.6,red!30!black,fill=red!40!black!10,rounded corners=1,
top color=red!40!black!20,bottom color=red!40!black!10,shading angle=20]
\tikzstyle{rope}=[brown!70!black,line width=1.2,line cap=round] 

\tikzstyle{force}=[->,myred,thick,line cap=round]
\tikzstyle{Fproj}=[force,myred!40]

\newboolean{showforces}
\setboolean{showforces}{true}

\usetikzlibrary{datavisualization}
\usetikzlibrary{datavisualization.formats.functions}

\tikzset{
	state/.style ={ellipse, draw, minimum width = 0.7 cm}
}
\usepackage[justification=centering]{caption}
\pgfplotsset{compat=1.15}
\usepackage{mathrsfs}
\usetikzlibrary{hobby,arrows.meta,3d}
\usepackage[all]{xy}
\usepackage{tikz}
\usepackage{subfig}
\usepackage{graphicx}
\usepackage{xcolor, soul, todonotes}
\usepackage[hypertexnames=false,debug,linktocpage=true,hidelinks]{hyperref}

\hypersetup{
	colorlinks,
	linktoc=all,
	linkcolor={blue},
	citecolor={red},
	urlcolor={blue}
}
\tikzstyle{punkt}=[circle, fill=black, minimum size=1mm,inner sep=0pt, draw]

\newtheorem{Theorem}{Theorem}[section]

\newtheorem{Corollary}[Theorem]{Corollary}
\newtheorem{Proposition}[Theorem]{Proposition}
\newtheorem{Remark}[Theorem]{Remark}

\newtheorem{Example}[Theorem]{Example}

\newtheorem{Definition}[Theorem]{Definition}

\newtheorem{Observation}[Theorem]{Observation}
\newtheorem{Notation}[Theorem]{Notation}
\let\epsilon\varepsilon
\let\kappa=\varkappa
\def\pnt{{\raise0.5mm\hbox{\large\bf.}}}

\newif\ifistoreview
\istoreviewtrue

\newcommand{\PIV}{\mathcal{P}\mathcal{I}\mathcal{V}}
\newcommand{\PIEV}{\mathcal{P}\mathcal{I}\mathcal{E}\mathcal{V}}
\newcommand{\SPIV}{\mathcal{S}\mathcal{P}\mathcal{I}\mathcal{V}}
\newcommand{\APIV}{\mathcal{A}\mathcal{P}\mathcal{I}\mathcal{V}}
\newcommand{\B}{\mathbb{B}}
\newcommand{\U}{\bm{\mathrm{U}}}
\renewcommand{\u}{\bm{\mathrm{u}}}
\newcommand{\SPACE}{\mathbb{S}\mathbb{P}\mathbb{A}\mathbb{C}\mathbb{E}}
\newcommand{\APACE}{\mathbb{A}\mathbb{P}\mathbb{A}\mathbb{C}\mathbb{E}}
\newcommand{\PINAC}{\mathcal{P}\mathcal{I}\mathcal{N}\mathcal{A}\mathcal{C}}
\newcommand{\PIENAC}{\mathcal{P}\mathcal{I}\mathcal{E}\mathcal{N}\mathcal{A}\mathcal{C}}
\newcommand{\APINAC}{\mathcal{A}\mathcal{P}\mathcal{I}\mathcal{N}\mathcal{A}\mathcal{C}}
\newcommand{\SPINAC}{\mathcal{S}\mathcal{P}\mathcal{I}\mathcal{N}\mathcal{A}\mathcal{C}}

\newcommand{\ACE}{\mathrm{ACE}}
\newcommand{\ACDE}{\mathrm{ACDE}}

\newcommand{\Do}{\mathrm{do}}
\newcommand{\h}{\mathbb{H}}
\newcommand{\I}{\mathbb{I}}
\newcommand{\kl}{\mathrm{KL}}
\newcommand{\Supp}{\mathrm{Supp}}
\newcommand{\PACE}{\mathbb{P}\mathbb{A}\mathbb{C}\mathbb{E}}

\renewcommand{\P}{\mathbb{P}}

\renewcommand{\phi}{\varphi}
\newcommand{\x}{\mathbf{x}}

\newcommand{\W}{\mathbf{W}}

\newcommand{\PEACE}{\mathbb{P}\mathbb{E}\mathbb{A}\mathbb{C}\mathbb{E}}

\newcommand{\Z}{\mathbf{Z}}
\newcommand{\z}{\mathbf{z}}
\newcommand{\E}{\mathbb{E}}

\begin{document}

	\title{Probabilistic Variational Causal Approach in Observational Studies}
\author[Usef Faghihi]{Usef Faghihi\,*}
\address{Usef Faghihi, Department of Mathematics and Computer Science, The University of Quebec at Trois-Rivieres, 3351 Bd des Forges, Trois-Rivières, QC G8Z 4M3, Canada.}
 \email{usef.faghihi@uqtr.ca}
 \thanks{\,*Corresponding author,  \texttt{usef.faghihi@uqtr.ca}}
\author[Amir Saki]{Amir Saki}
\address{Amir Saki, Department of Mathematics and Computer Science, The University of Quebec at Trois-Rivieres, 3351 Bd des Forges, Trois-Rivières, QC G8Z 4M3, Canada.}
\email{amir.saki@uqtr.ca, amir.saki.math@gmail.com}
	
\begin{abstract}
	In this paper, we introduce a new causal methodology that accounts for the rarity and frequency of events in observational studies based on their relevance to the underlying problem. Specifically, we propose a direct causal effect metric called the \textbf{P}robabilistic v\textbf{A}riational \textbf{C}ausal \textbf{E}ffect (PACE) and its variations adhering to certain postulates applicable to both non-binary and binary treatments. The PACE metric is derived by integrating the concept of total variation—representing the purely causal component—with interventions on the treatment value, combined with the probabilities of hypothetical transitioning between treatment levels.
	PACE features a parameter \( d \), where lower values of \( d \) correspond to scenarios emphasizing rare treatment values, while higher values of \( d \) focus on situations where the causal impact of more frequent treatment levels is more relevant. Thus, instead of a single causal effect value, we provide a causal effect function of the degree \( d \).
	Additionally, we introduce positive and negative PACE to measure the respective positive and negative causal changes in the outcome as exposure values shift. We also consider normalized versions of PACE, referred to MEAN PACE.  Furthermore, we provide an identifiability criterion for PACE to handle counterfactual challenges in observational studies, and we define several generalizations of our methodology. Lastly, we compare our framework with other well-known causal frameworks through the analysis of various examples.
\end{abstract}




\maketitle	


\section{Introduction}
As human beings, from the moment we wake up, we instinctively, whether consciously or unconsciously, try to identify what might be wrong. For example, why do I feel depressed? Is it because I didn’t sleep well last night due to health issues, or perhaps I overate and upset my stomach? These kinds of questions are inherently causal. Causal reasoning is the general, often an informal process of thinking about cause-and-effect relationships. Rooted in philosophy and cognitive science, it involves abstract, qualitative thinking without relying on formal methods. In daily life, people often causally reason about different subjects. Causal reasoning plays a role in nearly every aspect of our lives, from healthcare to industry and education. A scientific approach to causal reasoning is called causal inference. More precisely,``causal inference refers to an intellectual discipline that considers the assumptions, study designs, and estimation strategies that allow researchers to draw causal conclusions based on data''(see  \cite{hill2015causal}).

The most notable frameworks in causal inference are the Neyman-Rubin framework \cite{imbens2015, rubin1974} and the Pearl framework \cite{pearl2009causality, pearl2018}. Roughly speaking, the Neyman-Rubin framework is based on the idea of \textit{potential outcomes}. For instance, for an individual (or unit), a binary treatment \(T\) results in two potential outcomes that cannot be simultaneously observed: 1) the outcome when \(T = 0\), and 2) the outcome when \(T = 1\). The potential outcome that is not observable is called the \textit{counterfactual} outcome. The difference between these potential outcomes is defined as the \textit{individual causal effect} of \(T\) on the outcome variable. To aggregate the effects for the whole population, various ideas and formulas could be used (e.g., the expected value of the individual causal effects). To measure the \textit{direct causal effect} of \(T\) on the outcome variable, researchers employ formulas such as the \textit{controlled direct effect} and the \textit{natural direct effect}. The fundamental challenge in the Neyman-Rubin framework is the issue of missing data due to counterfactual outcomes. To address this, \textit{identifiability} criteria are employed. 

The Pearl framework is more general than the Neyman-Rubin framework and is based on the concept of \textit{intervention} in graphical causal models. Under certain assumptions, such as consistency and \textit{ignorability}/\textit{sufficiency}, an intervention where $T = t$ on average provides the same result as the potential outcome corresponding to $T = t$ (see \cite[Chapter~3]{pearl2009causality}). Pearl's framework utilizes \textit{Bayesian networks}, particularly through the use of Directed Acyclic Graphs (DAGs), to represent causal structures, allowing for a more comprehensive analysis of causality. 

Among the other causal frameworks, the information theoretic frameworks (\cite{janzing2013}) including the Janzing et al., the mutual information, and the conditional mutual information frameworks are notable. In the mutual information and conditional mutual information frameworks, the metrics \(\I(X;Y)\) and \(\I(X;Y \mid \Z)\) are used to quantify the causal strength of the arrow \(X \to Y\), where \(\Z\) represents all variables influencing \(Y\) other than \(X\) (i.e., the other parents of \(Y\)).
Janzing et al. introduced their causal framework based on five key assumptions for a DAG \(H\). These assumptions involve the local Markov condition, the relationship between causal strength and mutual information, and how causal strength is affected by the parents of nodes. The causal strength of an arrow \(X \to Y\) is interpreted as the effect of cutting the wire connecting nodes, where cutting alters the joint probability distribution. For a set of arrows \(\mathcal{S}\), the authors define the modified distribution \(\P_{\mathcal{S}}\), and the causal strength of \(\mathcal{S}\) in \(H\) is given by the Kullback-Leibler divergence \(D_{\mathrm{KL}}(\P || \P_{\mathcal{S}})\).
For further details on this framework, see Appendix \ref{janz}.

To elucidate the core principles of our causal methodology, consider an example involving the causal relationship between the presence of a symptom and the onset of a disease. Let \( X \) represent a random variable corresponding to three distinct levels of this symptom, where \( X \in \{x_1, x_2, x_3\} \). Additionally, let \( \Z \) denote other variables influencing the disease's presence. For a particular \( \Z = \z \), assume that the probability of observing \( x_1 \) is \( \mathbb{P}(x_1 \mid \z) \). If we could hypothetically revisit the moment before observing any level of the symptom and instead assume \( x_2 \) occurred, the probability of observing \( x_2 \) would be \( \mathbb{P}(x_2 \mid \z) \), with a similar statement holding for \( x_3 \).
Thus, in the hypothetical scenario where \( x_2 \) is observed, we adjust the likelihood of observing each level of the symptom according to its true probability, ensuring an accurate representation of reality in our analysis. By exploring 'what would have happened if \( x_2 \) occurred,' we delve into a central aspect of the Neyman-Rubin causal framework. Crucially, we do not modify the probabilities to make the likelihood of observing each level of symptom equal; rather, we retain their natural values.
This consideration is essential for our objective, which has two parts. First, we aim to derive a (direct) causal effect metric applicable to non-binary treatments. Second, we prioritize connecting the disease to the more common levels of the symptom, without giving equal weight to rare levels. Significant causal effects from rare levels of the symptom could disproportionately influence the overall causal effect of the symptom variable on the disease. Therefore, any causal change in the disease due to the hypothetical intervention from \( x_i \) to \( x_j \) must be weighted according to their likelihood of occurrence. This ensures the analysis reflects the frequency differences of each symptom level appropriately.
However, after an initial study of the relationship between the disease presence and the more common levels of the symptom, it is important to study the impact of the less common levels of the symptom as well.
In this scenario, the effect of hypothetically transitioning from a rare symptom \( x_i \) to \( x_j \) should not be weighted as insignificantly as the small probability of observing \( x_i \). Consequently, depending on the aim of the study or its sub-studies, rare values of the treatment variable (here, the symptom variable) may or may not hold importance.
Hence, to accommodate and adjust for the differences in symptom frequency—whether rare or common—and based on their relevance, we introduce a degree \( d \). By changing this degree \( d \), the importance of rare or frequent transitions between different levels of the symptom is adjusted, allowing for a clearer understanding of this impact.

In our methodology, the aforementioned weights corresponding to the (hypothetical) interventional transition from \( x_i \) to \( x_j \), integrated with the degree \( d \), are referred to as the \uline{natural availability of changing}  the exposure/treatment value of degree \( d \). For another example, assume that we are going to measure the direct causal effect of  the presence of a  natural noise $N$  on the quality $Q$ of a certain type of photo. Also, assume that $Q=g(N,\bm{\mathrm{O}})$, where $\bm{\mathrm{O}}$ denotes all other variables directly affecting $Q$. Also, assume that for each $\bm{\mathrm{O}}=\bm{\mathrm{o}}$, $N=1$ is very rare. Then, in real world data $\P(N=1|\bm{\mathrm{o}})$ is a very low value.  Hence,  this noise affects a photo  rarely. Thus, we need to incorporate $\P(N=1|\bm{\mathrm{o}})$ to calculate the direct causal effect of $N$ on $Q$, since low values for $\P(N=1|\bm{\mathrm{o}})$ means lower noisy photos. Therefore, our sense of causality is built by an integration of  the interventions $N=0$ and $N=1$, along with the natural availability of changing $N=0$ to $N=1$ while keeping $\bm{\mathrm{O}}=\bm{\mathrm{o}}$. However, if we consider the Neyman-Rubin/Pearl framework, the interventions $N=0$ and $N=1$ do not have anything to do with $\P(N=1|\bm{\mathrm{o}})$ (see Section~\ref{Comparing our Framework to the Pearl Framework}).  Consequently, we believe there is a crucial need for a generic causal methodology that can be integrated with the Neyman-Rubin/Pearl framework to address various types of causal questions.

Let $X$ be a random variable that we are looking for its direct causal effect on the outcome variable $Y$. Also, let $Y=g(X,\Z)$, where $\Z$ is the random vector consisting of all other variables directly affecting $Y$. In general, we assume that the above equation is the last equation of an \textit{structural equation model} (SEM) (see Section \ref{SEMdefinition}).  In this paper, as a part of our methodology, we provide a direct causal effect  metric called  probabilistic variational causal effect (PACE)  satisfying some  ideas and postulates (see Section~\ref{Ideas and Postulates}).   PACE is defined as an integration of the \textit{total variation} of a function (see Section~\ref{Total Variation})  and probability theory. The probabilistic part of the PACE of $X$ on $Y$ is the natural availability of changing  $X$ values keeping $\Z$ constant. PACE has a parameter $d$ determining the degree for which we consider the natural availability of changing  $X$ values. Lower values of \( d \) correspond to scenarios where rare cases are important. In contrast, higher values of \( d \) prioritize more frequent treatment values.  Hence, rather than providing a single value for the direct causal effect, we offer a direct causal effect function that depends on the degree \( d \). 
Also, we define some variations of PACE  and investigate their properties (see Section~\ref{PEACE, SPACE and APACE}). Additionally, we provide the positive and negative PACEs to measure the positive and negative causal changes in \( Y \) by incrementally changing the values of \( X \) while keeping \( \Z \) constant. In fact, if we change the value of \( X \) from \( x \) to \( x' \) (with \( x < x' \)) while keeping \( \Z \) constant, and observe an increase in \( Y \), we refer to this as a positive direct causal effect on \( Y \). Similarly, a negative direct causal effect is defined when \( Y \) decreases under the same conditions.
Furthermore, as required for observational studies, we provide an identifiability criterion for PACE (see Section~\ref{Identifiability of our Variational DCE Formulas}). 
We define MEAN PACE as a normalized version of PACE, based on the weights that appear in the PACE metric. In general, if for a given \(\Z = \z\), the MEAN PACE from terms associated with more frequent (hypothetical) transitions between treatment levels outweighs the other terms, then MEAN PACE for \(\Z = \z\) increases as the degree \(d\) increases. Otherwise, it decreases.
We also explain how PACE can be used to calculate the causal effect of a random variable \( W \) on \( Y \), even when \( W \) is not a direct cause of \( Y \), by considering specific paths from \( W \) to \( Y \). The main idea is to replace the \textit{mediators} with their parents and update the SEM (see Appendix  \ref{indirect}). In Appendix  \ref{matrix}, we go further and give  matrix representations for PACE and its variations, and then we extend this idea to define more general variational direct effect metrics.
In our variational causal metrics , we consider both the rarity and frequency of treatment values. However, in certain cases, a small segment of the population can heavily influence common causal metrics.  In Section \ref{general}, to regulate the impact of these rare or frequent subpopulations, we propose a generalization of our DCE metrics. Additionally, we explore some extended concepts regarding the natural availability of changing treatment values.

Our methodology has been applied to two\footnote{A third application in the area of economics, focusing on the renewal of a subscription, has been studied at \href{https://medium.com/@usef.faghihi/introducing-probabilistic-easy-variational-causal-effect-a-fresh-perspective-on-causal-inference-c8474dd8b5a6}{this link}.}  different areas: 
\begin{enumerate}
\item \textbf{Reinforcement Learning:} Deep Q Networks (DQN) have achieved success in various reinforcement learning tasks but are often hindered by spurious correlations due to their reliance on associative learning. In \cite{khelifi2024causal}, a novel approach integrates causal principles into DQNs using PACE. Incorporating PACE enhances the DQN’s ability to understand the environment's underlying causal structure, reducing the impact of confounding factors. Experimental results demonstrate that this new method outperforms standard DQNs, highlighting the effectiveness of causal inference via PACE in reinforcement learning. 
\item 
\textbf{Medical Sciences: }
In \cite{saki2024differentiating}, a novel approach was introduced to differentiate Gliosarcoma (GSM), a rare brain tumor, from Glioblastoma (GBM), a more common and aggressive brain tumor, by combining the Probabilistic Easy Variational Causal Effect (PEACE), a variation of PACE,  with the XGBoost algorithm. Unlike traditional methods that reduce dataset dimensions before causal analysis, the approach used in the aforementioned paper utilizes the full dataset, allowing for a detailed measurement of direct causal effects. This method enhances the understanding of causal relationships within the (high dimensional) data, providing more accurate differentiation between GSM and GBM.
\end{enumerate}

Now, we outline the structure of the paper. In Section \ref{Preliminaries}, we introduce foundational concepts, including total variation, directed acyclic graphs, the Neyman-Rubin and Pearl frameworks, and structural equation models. In Section \ref{Probabilistic Variational Causal Effect Framework}, we present our methodology, providing the PACE metric and discussing its properties. Additionally, we introduce variations of PACE, namely PEACE, SPACE, and APACE, and explore their characteristics and interrelationships. Positive and negative versions of our direct causal metrics are also introduced and analyzed. The identifiability of PACE and its variations is examined as well. Section \ref{The MEAN Versions of PEACE, PACE and APACE} focuses on the mean versions of our direct causal metrics and their properties. In Section \ref{simpson}, we analyze an example of Simpson's paradox with a non-binary treatment variable using our approach. In Section \ref{Comparing our Framework to the Pearl Framework}, we compare our methodology to the Neyman-Rubin/Pearl framework, highlighting similarities and differences. Section \ref{Investigating Some Examples} investigates three examples to demonstrate the generalization capacity of PACE. Finally, in Section \ref{general}, we introduce several generalizations of our methodology from different perspectives.

This paper also includes several appendices. In  Appendix   \ref{janz}, we provide details on the framework proposed by Janzing et al. Appendix  \ref{indirect} is dedicated to calculating indirect causal effects using our methodology.  Appendix  \ref{matrix} presents a matrix representation for each of our causal metrics and introduces a general method for using matrix representations to define new causal metrics.  Appendix  \ref{Proofs of Results} contains the proofs of the theorems and propositions presented throughout the paper. Finally,  Appendix  \ref{different} provides examples that illustrate certain properties discussed in the main text.

In this paper, all the random variables discussed are finite unless otherwise stated. 

\section{Preliminaries}\label{Preliminaries}
In this section, we present the foundational concepts, covering total variation, directed acyclic graphs, the Neyman-Rubin and Pearl frameworks, as well as structural equation models.
\subsection{Total Variation}\label{Total Variation}
In this subsection, we briefly explain the total variation formula for functions of one variable and sequences of real numbers. Further,  the definitions of the total positive/negative variations  of a function and a sequence are provided.

\subsubsection{Total Variation of a Function of one Variable}
Let $f:[a,b]\to\mathbb{R}$ be a function. A \textit{partition} for the interval $[a,b]$ is a an ordered set $P=\{a=x_0^{(P)}<x_1^{(P)}<\cdots <x_{n_P-1}^{(P)}<x_{n_P}^{(P)}=b\}$. Denote the set of all partitions of $[a,b]$ by $\mathcal{P}([a,b])$. Then, the total variation of $f$ is defined as follows:
\[\mathcal{V}(f):=\sup_{P\in\mathcal{P}([a,b])}\sum_{i=1}^{n_P}|f(x_{i}^{(P)})-f(x_{i-1}^{(P)})|.\]
To the best of our knowledge, the idea of the total variation for the first time was discussed in \cite{Jordan1881}.

It is well-known that if $f$ is continuously differentiable, then $\mathcal{V}(f)=\int_a^b|f'(t)|\,\mathrm{d}t$.  This integral represents the arc length of the curve described by \( f(t) \) from \( t = a \) to \( t = b \) in \(\mathbb{R}\). Therefore, \( \mathcal{V}(f) \) can be interpreted as the distance a particle travels along this curve when its position at time \( t \), within the interval \([a, b]\), is given by \( f(t) \).

\subsubsection{Total Variation of a Sequence of Real Numbers}\label{Total Variation of a Sequence of Real Numbers}
Let $\{x_n\}_{n=0}^{\infty}$ be a sequence of real numbers. Then, the total variation of this sequence is defined as $\mathcal{V}\left(\{x_n\}_{n=0}^{\infty}\right)=\sum_{n=1}^{\infty}|x_n - x_{n-1}|$ (see \cite[Chapter IV]{Dunford1958}).
\subsubsection{Total Positive and Negative  Variations}

Let $r\in\mathbb{R}$. Define $r^{\bm{+}}:=\max\{r,0\}$ and $r^{\bm{-}}:=|r|-r^{\bm{+}}$. Then, the total positive variation of a function $f:[a,b]\to\mathbb{R}$ is defined as follows (see \cite{Jordan1881}):
\[\mathcal{V}(f)^{\bm{+}}:=\sup_{P\in\mathcal{P}([a,b])}\sum_{i=1}^{n_P}\left(f(x_i^{(P)})-f(x_{i-1}^{(P)})\right)^{\bm{+}}.\]
Similarly, the total  negative variation of $f$ is defined as follows:
\[\mathcal{V}(f)^{\bm{-}}:=\sup_{P\in\mathcal{P}([a,b])}\sum_{i=1}^{n_P}\left(f(x_i^{(P)})-f(x_{i-1}^{(P)})\right)^{\bm{-}}.\]
One could easily see that 
$\mathcal{V}(f) = \mathcal{V}(f)^{\bm{+}} + \mathcal{V}(f)^{\bm{-}}$.
The total positive and negative variations could be similarly defined for  sequences of real numbers. In this case, one could see that $\mathcal{V}\left(\left\{x_n\right\}_{n=0}^{\infty}\right) = \mathcal{V}\left(\left\{x_n\right\}_{n=0}^{\infty}\right)^{\bm{+}} + \mathcal{V}\left(\left\{x_n\right\}_{n=0}^{\infty}\right)^{\bm{-}}$.



\subsection{Directed Acyclic Graphs}\label{Directed Acyclic Graphs}
A directed acyclic graph (DAG) is a directed graph without any directed cycles. Let $H$ be a DAG, and $v$ be a node of $H$. Then, a node $u$ is called a \textit{parent} of $v$ if there exists a directed edge $u\to v$. An \textit{ancestor} of $v$ in $H$ is a node $w$ of $H$ that is connected to $v$ by a directed path originating from  $w$. \textit{Children} and \textit{descendants} of nodes in $H$ are defined similarly. 

\subsection{Neyman-Rubin Framework
}
Consider calculating the causal effect of a binary random variable \( S \) (e.g., smoking) on another random variable \( C \) (e.g., the lung canser). When \( S = 1 \) for a particular \textit{unit}, the value of \( C \) is observed. However, for\( S = 0 \), the value of \( C \) is missing, creating what is known as a \textit{counterfactual}. Counterfactuals present a significant challenge in causal inference. To address this issue, researchers employ identifiability criteria to transform the causal question into a statistical one, as discussed in Section \ref{Identifiability of our Variational DCE Formulas}.
The Neyman-Rubin framework examines three outcome variables: 1) \( C \), the \textit{observed outcome}; 2) \( C(0) \), the \textit{potential outcome} when \( S = 0 \); and 3) \( C(1) \), the \textit{potential outcome} when \( S = 1 \). In a study, only one of the potential outcomes is observed at any given time.
Another important consideration in causal inference is to account for \textit{confounders}, which are assumed to be the common cause of $S$ and $C$ at the same time (for instance, in a clinical study, the patient condition is a confounder for both treatment and outcome).  Note that different causal effect formulas are commonly used in the Neyman-Rubin framework such as the followings:
\begin{itemize}
\item 
Average Treatment Effect (ATE) or Average Causal Effect (ACE) of $S$ on $C$ is defined as 
$\E(C(1)-C(0))$.
\item 
The Conditional Average Treatment Effect (CATE) or Conditional Average Causal Effect (CACE) of \( S \) on \( C \), given \( X = x \), is denoted by \( \tau(x) \) and defined as:
$ \tau(x) = \E(C(1) - C(0) \mid X = x)$, 
where \( X \) is a random variable that specifies different subpopulations and is technically referred to as a \textit{covariate}. It can be shown that the ACE of \( S \) on \( C \) is the expected value of CACE over \( X \). More precisely:
\[ \E(C(1) - C(0)) = \E_X\left(\E(C(1) - C(0) \mid X = x)\right) = \E(\tau(X)). \]
\item 
Average Controlled Direct Effect (ACDE) of $S$ on $C$, controlling $M$, is defined  as: \[\ACDE\bigl((S\to C)|M\bigr)=\mathbb{E}_M\left(C(S=1, M=m)-C(S=0, M=m)\right),\]
where $M$ denotes a subset of all  other endogenous variables, and $C(S=s, M=m)$ means the potential outcome of $C$ with respect to $S=s$ and $M=m$.
\item 
Average Natural Direct Effect (ANDE) of $S$ on $C$ by setting $G=g$ is defined as follows:
\[\E\Bigl(C(S=1,M=M(S=0))-C(S=0,M=M(S=0))\Bigr),\]
where $M$ is such as the above, and  \( G \) represents all the parents of \( S \) excluding \( C \).
\end{itemize}

\subsection{Pearl Framework}\label{Pearl Model}

Pearl uses DAGs equipped with the \textit{local Markov assumption} to simulate and study causal relationships between variables. The local Markov assumption states that a node, given its parents, is independent of all its non-descendants. This implies that if we denote the treatment by \( S \), the outcome by \( C \), and let \( G \) represent all parents of \( C \) excluding \( S \), the following decomposition holds:
\[
\P(c, g, s) = \P(g) \P(s \mid g) \P(c \mid g, s).
\]
Using probability theory, the local Markov assumption, and by intervening in the DAGs’ nodes, one can estimate the effect(s) of possible cause(s) of the events \cite{pearl2018book}. Similar to the Neyman-Rubin framework, Pearl \cite{pearl2018book} computes ACE by subtracting the means of the treatment and control groups.  Therefore,
\[
\ACE(S \to C) = \E(C \mid \Do(S=1)) - \E(C \mid \Do(S=0)),
\]
where \(\Do(S=s)\) means intervening on \(S\), resulting in a constant value \(S=s\) for the whole population. ACE in the Pearl framework, under the consistency and sufficiency assumptions\footnote{The sufficiency assumption ensures that all confounders are present in the model, which aligns with the back-door criterion in Pearl's framework, allowing for proper adjustment.}, is equal to the ACE in the Neyman-Rubin framework. The CACE and ACDE formulas defined in the Neyman-Rubin framework can be restated in the Pearl framework (under the same assumptions) as follows:
\begin{align*}
&\mathrm{CACE}(S\to X\mid X)=\E(C|\mathrm{do}(S=1),X=x)-\E(C|\mathrm{do}(S=0),X=x)\\
&\ACDE(S\to C\mid M)=\E_G(C|\mathrm{do}(S=1, M=m))-\E(C|\mathrm{do}(S=0, M=m)),
\end{align*}
where $M$ is a subset of $G$.
Although the ANDE formula is used in the Pearl framework as well, it does not have a do-expression.

Now, we briefly explain how to calculate the ACE of $S$ on $C$ in the Pearl framework. To do so, we have that
\[\P(c)=\sum_{s,g}\P(c,s,g)=\sum_{s,g}\P(g)\P(s|g)\P(c|g,s).\]
After the intervention $S=s_0$, the post probability value $\P(s|g)$  turns into $1$ for $s=s_0$, and 0 otherwise  (graphically, this means removing the causal arrow from $G$ pointing to $S$, and it is called the \textit{modularity assumption}). Thus, we have that
\[\P(c|\Do(S=s_0))=\sum_g\P(g)\P(c|g,s_0).\]


\subsection{Structural Equation Model}\label{SEMdefinition} A
structural equation model (SEM) consists of: 
\begin{enumerate}
\item
a set of \textit{endogenous variables}, namely the variables that we are interested in their causes,
\item
a set of \textit{exogenous variables}, namely the variables that we are not interested in their causes but interested in them when they are causes of endogenous variables, and
\item
a set of functional relationships $Y=f(X_1,\ldots, X_n, Z_1,\ldots, Z_m)$, where $Y$, and $X_1,\ldots, X_n$ are endogenous variables, and $Z_1,\ldots, Z_m$ ‌are exogenous variables. Here, each of $X_1,\ldots, X_n$ and $Z_1,\ldots, Z_m$ is called a \textit{direct possible cause} of $Y$.
\end{enumerate}
An example of a linear SEM is as follows:
\begin{align*}
X=U_X,\qquad 
Y=\alpha X+ U_Y,\qquad
Z= \beta Y + \gamma X +U_Z,
\end{align*}
where $U_X, U_Y$ and $U_Z$ are exogenous variables, and $X, Y$ and $Z$ are endogenous variables.

SEMs  were initially studied by Sewall Wright to causally investigate regression equations in genetics (see \cite{wright1920} and \cite{wright1921}). Since then, several diverse methods of SEMs have been suggested in different branches of science, business, psychology, social sciences, etc (see \cite{bollen2013}, \cite{boslaugh2007}, \cite{duncan2014}, and
\cite{english2006}). 

\section{Probabilistic Variational Causal Effect Framework}\label{Probabilistic Variational Causal Effect Framework}
In this section, we introduce our methodology, presenting the PACE metric and discussing its key properties. Furthermore, we propose several variations of PACE—namely PEACE, SPACE, and APACE—and examine their characteristics and interrelationships. Both positive and negative versions of our direct causal metrics are also introduced and analyzed. Additionally, we investigate the identifiability of PACE and its variations.
\subsection{Interventions in SEMs}\label{Interventions in SEMs}
Let $X,X_1\ldots, X_n$ and $Y$ be random variables and $Y=g(X,\bm{\mathrm{Z}})$, where we have that $\Z=(X_1,\ldots, X_n)$. Suppose that we are asked to calculate the \underline{direct causal effect} of $X$ on $Y$.  Here, $X$  and $\bm{\mathrm{Z}}$ are not necessarily independent or causally unrelated.  Assume that there is a functional relationship $X=h(\bm{\mathrm{Z}'}, \bm{\mathrm{W}})$, where $\bm{\mathrm{W}}=(W_1,\ldots, W_m)$ and $\bm{\mathrm{Z}}'=(X_{i_1},\ldots, X_{i_k})$ with $1\le i_1<\cdots <i_k\le n$. By the intervention $X=x_0$, we mean setting $X=x_0$ and replacing the relationship $X=h(\bm{\mathrm{Z}'} , \bm{\mathrm{W}})$ with $X=x_0$.
\begin{Notation}
After the intervention $X=x_0$, we denote the (potential) outcome by $Y_{x_0} = Y(X=x_0)= g_{in}^X(x_0,\bm{\mathrm{Z}})$, where $g_{in}^X(x_0,\Z)$ is $g(X,\Z)$ that we replace any functional relationship defining $X$ with $X=x_0$. Now, assume that we have the intervention $(X,\Z)=(x_0,\z)$. Then, for simplicity, we denote $g_{in}^{(X,\Z)}(x_0,\z)$ by $g_{in}(x_0,\z)$.
\end{Notation}
In terms of SEMs, assume that we are given the following SEM:
\[
\begin{array}{l}
\left.\begin{array}{l} \text{Equation } (1)\\ \hspace*{1cm}\vdots \\ \text{Equation }(m+k)\end{array}\right\}\text{The equations defining the variables $\bm{\mathrm{W}}$ and $\bm{\mathrm{Z}}'$.}\\\vspace*{-0.2cm}\\
\hspace*{0.2cm}X=h(\bm{\mathrm{Z}'} , \bm{\mathrm{W}})\\[2pt]
\left.\begin{array}{l} \text{Equation } (m+k+2)\\ \hspace*{1cm}\vdots \\ \text{Equation }(m+n+1)\end{array}\right\}\begin{array}{l}\text{The equations defining the variables}\\ \text{in $\bm{\mathrm{Z}}$ other than  $\bm{\mathrm{Z}}'$.}\end{array}\\\vspace*{-0.2cm}\\
\hspace*{0.23cm}Y=g(X,\bm{\mathrm{Z}})
\end{array}
\]
After the intervention $X=x_0$, we have a new SEM which coincides with the above SEM except for the functional relationship defining $X$, which is $X=x_0$ instead of $X=h(\bm{\mathrm{Z}'} , \bm{\mathrm{W}})$. 
Note that intervening and conditioning are two different concepts since conditioning on $X=x_0$ leaves the equality $x_0=h(\bm{\mathrm{Z}'} , \bm{\mathrm{W}})$, which makes $\bm{\mathrm{Z}}'$ and $\bm{\mathrm{W}}$ dependent to each other.

\subsection{Ideas and Postulates}\label{Ideas and Postulates}
To formalize the direct causal effect (DCE) of \( X \) on \( Y \), our approach concentrates on the total potential causal variations in \( Y \) in response to interventional modifications in \( X \) considering the natural availabilities of these changes. Subsequently, as we will see, an aggregation method, such as summation or determining the maximum, can be utilized to quantify these causal changes.

We start with the following general \underline{ideas}:
\begin{enumerate}[label=\textbf{I.\arabic*}]
\item \label{stochastic nature}
The stochastic nature of variables should be considered in our DCE metric.
\item \label{direct causal effect}
Our metric of causal effect becomes a direct causal effect metric in the following sense:
measuring the changes in $Y$, by changing the value of $X$ and  keeping $\bm{\mathrm{Z}}$ constant.
\item \label{restrictions}
Our DCE should be applicable for different restrictions of $X$\footnote{Let $X:\Omega\to\mathbb{R}$ be a random variable, and $P\subseteq\mathbb{R}$. Then, we define the restriction of $X$ on $P$, denoted by $X_P$, to be the function $X_P:X^{-1}(P)\to\mathbb{R}$.}  that are not necessarily random variables by a standard definition of random variables (the sum of probability values of a restriction of $X$ could be less than 1). Further, we expect that the DCE of a restriction of $X$ on $Y$ is not greater than the DCE of $X$ on $Y$.
\item \label{comparing}
If a different functional relationship \( Y' = \tilde{g}(X, \bm{\mathrm{Z}}) \) existed such that variations in \( X \), while keeping \( \bm{\mathrm{Z}} \) constant, led to smaller fluctuations in \( Y' \) compared to \( Y \), then the DCE of \( X \) on \( Y' \) would be smaller than the DCE of \( X \) on \( Y \).

\item\label{degree}
Our DCE admits a degree $d\ge 0$  to control, augment or reduce the effects of probability values.  Because in some real-world problems, rare cases are important, while in others, they are not. Here, we describe the choices $d=0$ and $d=1$ for our DCE to be \textit{non-probabilistic} and \textit{commonly probabilistic}, respectively. 
\end{enumerate}
Building on \ref{direct causal effect}, since we can hold \( \bm{\mathrm{Z}} \) constant at any value within its support, the DCE must incorporate an aggregation method, such as the expected value, to accommodate all possible values of \( \bm{\mathrm{Z}} \).
In \ref{direct causal effect} and \ref{comparing} , by ``possible interventional changes in $Y$ values (with respect to changing $X=x$ to $X=x'$ while keeping $\Z=\z$)'', we mean the value of $|g_{in}(x',\z)-g_{in}(x,\z)|$ with $\P(x'|\z),\P(x|\z)>0$.

Now, with the aforementioned ideas in mind, we introduce the following \underline{postulates} to define our DCE:

\begin{enumerate}[label=\textbf{P.\arabic*}]
\item \label{locality}
Given $Y=g(X,\bm{\mathrm{Z}})$ and $d\ge 0$, the DCE of degree $d$ of $X$ on $Y$  depends only on $g$, $d$, and the  joint probability distribution of $X$ and $\bm{\mathrm{Z}}$.
\item \label{restrction does not increase}
Removing  some values of $X$ does not increase the DCE of degree $d$ of $X$ on $Y$(by \ref{restrictions}).

\item \label{conditional DCE}
A controlled DCE of degree \(d\) of \(X\) on \(Y\), keeping \(\bm{\mathrm{Z}} = \bm{\mathrm{z}}\), can be defined such that the expected value of these controlled DCEs of degree \(d\) over \(\bm{\mathrm{Z}}\) aligns with our defined DCE of degree \(d\).
\item \label{causal sense}
For \(d > 0\), the controlled DCE of degree \(d\) of \(X\) on \(Y\), with \(\bm{\mathrm{Z}} = \bm{\mathrm{z}}\), equals zero if and only if the value of \(Y\) remains unchanged under any possible interventions on \(X\) while keeping \(\bm{\mathrm{Z}} = \bm{\mathrm{z}}\).

\item \label{binary DCE}
If $X$ is a binary random variable with $\mathrm{Supp}(X)=\{0,1\}$, then the controlled DCE of degree $d$ of $X$ on $Y$ keeping $\bm{\mathrm{Z}}=\bm{\mathrm{z}}$ equals $|g_{in}(1,\bm{\mathrm{z}})-g_{in}(0,\bm{\mathrm{z}})|w_d(X=1, X=0,\bm{\mathrm{z}})$, where $w_d(X=1, X=0,\bm{\mathrm{z}})$ encodes the natural availability of degree $d$ of changing $X$ from $X=0$ to $X=1$ keeping $\bm{\mathrm{Z}}=\bm{\mathrm{z}}$.
\end{enumerate}

We should make it clear what we mean by natural availability of changing in the $X$ value in \ref{binary DCE}. To do so, first we investigate a simple example. Assume that $Y=g(X,Z)=X+Z$ and $X=h(Z)$, where $X$ and $Z$ are binary random variables. Obviously, for a constant $Z=z$, we have that $|g_{in}(1,z)-g_{in}(0,z)|=1$, while  it is not possible to have any changes in $X$ values (because, $X$ is a function of $Z$). Hence, it is reasonable to have that $w_d(X=1, X=0,z)=0$. In such a situation, we say that the interventional change of $Y$ value with respect to changing $X=0$ to $X=1$ (or vice versa) while keeping $\Z=\z$,  is not possible. Therefore, in general,  by the notation in Section \ref{Interventions in SEMs}, the  relationship $X=h(\bm{\mathrm{Z}}', \bm{\mathrm{W}})$ and the stochastic natures of $X$ and $\bm{\mathrm{Z}}$ make some restrictions on interventional changes of $X$ keeping $\Z=\z$, which could be encoded in $w_d(X=1, X=0,\bm{\mathrm{z}})$.

\subsection{Probabilistic Variational Causal Effect Metric}	
In this subsection, we define a DCE metric satisfying the   postulates explained in Section~\ref{Ideas and Postulates}, and we call it the Probabilistic Variational Causal Effect metric (PACE). First, we define the PACE of degree $d=1$. For the sake of simplicity, we drop the phrase "degree 1" for what we define in the sequel (i.e., instead of the PACE of degree 1, we only say PACE). To satisfy \ref{conditional DCE}, we will naturally define PACE as the expected value of \textit{controlled PACE}. Hence, it is enough to define the controlled PACE in the sequel. 

Let us assume that $\mathrm{Supp}(X)=\{x_0,\ldots, x_l\}$\footnote{For a given random variable $X$, we define the support of $X$, denoted by $\Supp(X)$, as the set of all possible values of $X$ (i.e., the set of all values $X=x$ with $\P_X(x)>0$).} with $x_0<\cdots<x_l$. Then, we define the \textit{interventional variation} of $Y$ with respect to $X$ keeping $\Z=\z$, denoted by $\mathcal{I}\mathcal{V}^{\z}(X\to Y)$, to be the total variation of the sequence $\bigl(g_{in}(x_0,\z),\ldots, g_{in}(x_l,\z)\bigr)$ (see Section \ref{Total Variation of a Sequence of Real Numbers}). Thus, we have that
\[\mathcal{I}\mathcal{V}^{\z}(X\to Y)=\sum_{i=1}^l\bigl|g_{in}(x_i,\z)-g_{in}(x_{i-1},\z)\bigr|.\]


This formula is a starting point to define the controlled PACE. Note that the above formula is not probabilistic, and in a real world problem some values of $X$  could be more probable than  other ones. It follows that the natural availability of changing the value of \(X\) from \(x_{i-1}\) to \(x_i\) (possibly) as a function of the likelihood of \(x\) and \(x'\) differs for various pairs of values \((x_{i-1}, x_i)\).
Hence, we associate a weight (as it was stated in \ref{binary DCE}) to each  summand of the above formula as the natural availability of changing in $X$ values, keeping $\Z=\z$. \ul{To do so, for the $i^{\text{th}}$ summand, we can independently select $x_i$ and $x_{i+1}$ given $\Z=\z$, with the probability of
$\P(x_i|\z)\P(x_{i-1}|\z)$}. 
Thus, one may suggest defining the controlled PACE of $Y$ with respect to $X$ keeping $\Z=\z$  as follows:
\begin{equation}\label{piev}
\sum_{i=1}^{l}|g_{in}(x_{i},\z)-g_{in}(x_{i-1},\z)|\P(x_i|\z)\P(x_{i-1}|\z).
\end{equation}
However, this formula  is not still a suitable candidate for the controlled PACE, because its amount might increase if we remove some values of $X$\footnote{For example, let us assume that $\mathrm{Supp}(X)=\{x_0,x_1,x_2\}$ with $x_0<x_1<x_2$, and we have that $\P_X(x_0)=a, \P_X(x_1)=b$ and $\P_X(x_2)=c$. One could find an increasing function $Y=g(X)$ with $g_{in}(x_1)-g_{in}(x_0)=g_{in}(x_2)-g_{in}(x_1)=1$, and $2ac>ab+bc$.}, which contradicts our expectation of a DCE metric (it violates \ref{restrction does not increase}). Nevertheless, we call the above formula the \textit{probabilistic interventional easy variation} of $Y$ with respect to $X$ keeping $\Z=\z$, and we denote it by $\mathcal{P}\mathcal{I}\mathcal{E}\mathcal{V}^{\bm{z}}(X\to Y)$ (later, we will introduce a different DCE metric known as PEACE, such that \(\mathcal{P}\mathcal{I}\mathcal{E}\mathcal{V}^{\bm{z}}(X \rightarrow Y)\) represents the controlled PEACE).

Let us take a second look at the total variation of a sequence. Assume that $\{\alpha_n\}_{n=0}^{\infty}$ is a sequence of real numbers. Then, one could see that
\begin{align*}
\mathcal{V}\bigl(\{\alpha_n\}_{n=0}^{\infty}\bigr)&=
\sup\left\{\sum_{i=1}^{\infty}|\alpha_{n_i}-\alpha_{n_{i-1}}|: \{\alpha_{n_i}\}_{i=0}^{\infty} \text{ is a subsequence of } \{\alpha_n\}_{n=0}^{\infty}\right\}\\
&=	\sup\bigl\{\mathcal{V}\left(\{\alpha_{n_i}\}_{i=0}^{\infty}\right): \{\alpha_{n_i}\}_{i=0}^{\infty} \text{ is a subsequence of } \{\alpha_n\}_{n=0}^{\infty}\bigr\}.
\end{align*}
Let us return to our random variable \(X\). By an abuse of notation, we refer to a subsequence \(\bigl(x_0^{(P)}, \ldots, x_{n_P}^{(P)}\bigr)\) of \(\bigl(x_0, \ldots, x_l\bigr)\) as a \textit{partition} \(P\) for \(X\), and thus denote the restriction of \(X\) to the partition \(P\) by \(X_P\). We also denote the set of all partitions of \(X\) by \(\mathcal{P}(X)\). Building on the concept from the above formulation for the total variation of a sequence, we can define the \textit{probabilistic interventional variation} of \(Y\) with respect to \(X\), keeping \(Z = z\), as follows:
\[\mathcal{P}\mathcal{I}\mathcal{V}^{\bm{\z}}(X\to Y)=\sup\left\{\mathcal{P}\mathcal{I}\mathcal{E}\mathcal{V}^{\bm{z}}(X_P\to Y): P\in\mathcal{P}(X)\right\},\]
which implies that
\[\mathcal{P}\mathcal{I}\mathcal{V}^{\bm{\z}}(X\to Y)=\max_{P\in\mathcal{P}(X)}\sum_{i=1}^{n_P}\left|g_{in}(x_{i}^{(P)},\z)-g_{in}(x_{i-1}^{(P)},\z)\right|\P(x_i|\z)\P(x_{i-1}|\z).\]

Finally, we define the controlled PACE of $X$ on $Y$ keeping $\Z=\z$, to be $\mathcal{P}\mathcal{I}\mathcal{V}^{\z}(X\to Y)$.
Now, to formulate a DCE of $X$ on $Y$, we can simply take the expected value of $\mathcal{P}\mathcal{I}\mathcal{V}^{\z}(X\to Y)$ with respect to $\Z$. This approach for a DCE is termed the Probabilistic vAriational Causal Effect, denoted by \(\PACE(X \to Y)\). Hence, 
$\PACE(X\to Y):=\E_{\Z}\left(\mathcal{P}\mathcal{I}\mathcal{V}^{\z}(X\to Y)\right)$.
Especially, when  $X$ is binary, we have that 
\[\mathcal{P}\mathcal{I}\mathcal{V}^{\z}(X\to Y)=|g_{in}(1,\z)-g_{in}(0,\z)|\P(X=1|\z)\P(X=0|\z).\]
Consider the following example: we aim to calculate the direct causal effect of a rare disease on blood pressure, denoted by \(X=1\) for the presence of the disease and \(X=0\) otherwise. In this scenario, opting for \(\mathcal{I}\mathcal{V}^{z}(X \to Y)\) as the controlled DCE of \(X\) on \(Y\) keeping \(Z=z\) is more appropriate than using \(\mathcal{P}\mathcal{I}\mathcal{V}^{z}(X \to Y)\). This is because using \(\mathcal{P}\mathcal{I}\mathcal{V}^{z}(X \to Y)\) leads to an underestimate of the causal effect due to the rare occurrence of the disease (when \(X=1\)). Consequently, this low causal effect does not accurately reflect the causal significance of the rare disease on blood pressure.
This motivates us to define the PACE of degree $d$ for any $d\ge 0$. Indeed, we define the  probabilistic interventional variation of degree $d$ of $Y$ with respect to $X$ keeping $\Z=\z$, denoted by $\mathcal{P}\mathcal{I}\mathcal{V}_d^{\z}(X\to Y)$ as follows:
\begin{equation}\label{piv}
\mathcal{P}\mathcal{I}\mathcal{V}_d^{\z}(X\to Y):=\max_{P\in\mathcal{P}}\sum_{i=1}^{n_P}|g_{in}(x_{i}^{(P)},\z)-g_{in}(x_{i-1}^{(P)},\z)|\P(x_{i}^{(P)}|\z)^d\P(x_{i-1}^{(P)}|\z)^d.
\end{equation}
Finally, we define the probabilistic variational causal effect (PACE) of degree $d$ of $X$ on $Y$ as follows:
$	\PACE_d(X\to Y):=\E_{\Z}\left(\mathcal{P}\mathcal{I}\mathcal{V}_d^{\z}(X\to Y)\right)$.
\begin{Theorem}\label{checkingpostulatesforpace}
$\PACE_d(X\to Y)$ satisfies \ref{locality}, \ref{restrction does not increase}, \ref{conditional DCE}, \ref{causal sense}, and \ref{binary DCE}.
\end{Theorem}
\begin{proof}
See Appendix  \ref{3.2}.
\end{proof}
We have the following observation for binary random variables  $X$ and $Y$.
\begin{Observation}
Let $X$ and  $Y$ be binary random variables with $\Supp(Y)=\{0,1\}$. Then, $\PACE_{d}(X\to Y)$ is the $d^{\text{th}}$ (fractional) moment of the random variable $\mathcal{P}\mathcal{I}\mathcal{V}^{\z}(X\to Y)$ with respect to $\Z$, namely  
\[\PACE_{d}(X\to Y)=\E_{\Z}\left(\left(\mathcal{P}\mathcal{I}\mathcal{V}^{\z}(X\to Y)\right)^d\right).\]
\end{Observation}
In the following theorem, we provide another property of PACE for the composition of maps. 
\begin{Proposition}\label{prop1}
Let us assume that $X=h(W,\Z)$ and $Y=g(X,\Z)$. Hence, we have that $Y=\widetilde{g}(W,\Z)=g(h(W,\Z),\Z)$. Then, we have that $\PACE_d(W\to Y)=0$ if  $\PACE_d(X\to Y)=0$.
\end{Proposition}
\begin{proof}
See Appendix   \ref{3.4}.
\end{proof}

\subsection{Degree $d$}
To gain a better understanding of our PACE metric, we can investigate  $F(d)=\PACE_d(X\to Y)$ for $d\ge 0$. A discrete approach involves partitioning  $[0,1]$ such that $d_i=i/N$ for any $0\le i\le N$, where $N$ is a positive integer. To extend beyond this interval, consider larger intervals $[0,M]$, where $M$ is a positive real number with $M>1$ (in this case, if $d_i=\frac{iM}{N}$, then $\{d_0,\ldots, d_N\}$ is a partition for $[0,M]$). Then, we can define the \textit{PACE $N$-vector} of degree $d$ of $X$ on $Y$ as follows:
\[\bigl(\PACE_{d_0}(X\to Y),\ldots, \PACE_{d_N}(X\to Y)\bigr).\]
We note that \(\PACE_{d_0}(X \to Y)\) is non-probabilistic and particularly useful when \uline{rare cases are highly significant}. As \( d \) approaches 1, the influence of the probability values \( \P(X \mid \bm{Z}) \) on \(\PACE_d(X \to Y)\) increases, making \(\PACE_d(X \to Y)\) more relevant in contexts where rare cases are less significant.

\begin{Observation}
If $d_2\ge d_1\ge 0$, then 
\[\PACE_{d_2}(X\to Y)\le \PACE_{d_1}(X\to Y).\]
\end{Observation}
Now, we briefly explain an intuition behind the degree $d$ in our PACE metric. Let us assume that $d_2>d_1>0$. 
Also, assume that $1\le i ,j\le n_P$ are in such a way that $\bigl(\P(x_{i-1}^{(P)}|\z)\P(x_i^{(P)}|\z)\bigr)^{d_1}/\bigl(\P(x_{j-1}^{(P)}|\z)\P(x_j^{(P)}|\z)\bigr)^{d_1}>1$. Then, we have that
\begin{align*}
\left(\frac{\P(x_{i-1}^{(P)}|\z)\P(x_i|^{(P)}\z)}{\P(x_{j-1}^{(P)}|\z)\P(x_j^{(P)}|\z)} \right)^{d_2}> \left(\frac{\P(x_{i-1}^{(P)}|\z)\P(x_i^{(P)}|\z)}{\P(x_{j-1}^{(P)}|\z)\P(x_j|\z)}\right)^{d_1}
\end{align*}
if and only if
\[ \left(\frac{\P(x_{i-1}^{(P)}|\z)\P(x_i^{(P)}|\z)}{\P(x_{j-1}^{(P)}|\z)\P(x_j^{(P)}|\z)} \right)^{d_2-d_1}>1,
\]
while the latter holds. Hence, by increasing the degree $d$, the relative  effect of the weight $\bigl(\P(x_{i-1}^{(P)}|\z)\P(x_i^{(P)}|\z)\bigr)^{d}$ compared to $\bigl(\P(x_{j-1}^{(P)}|\z)\P(x_j^{(P)}|\z)\bigr)^{d}$ gets higher, and consequently it has a greater share in the value of $\PIEV_d^{\z}(X_P\to Y)$. \uline{It follows that by increasing $d$, the elements of $\mathrm{Supp}(X)$ with  higher probability values given $\Z=\z$ are more effective in calculating $\PIEV_d^{\z}(X_P\to Y)$.}
\begin{Remark}
If $\PIV_d^{\z}(X\to Y)=\PIEV_d(X_P\to Y)$ for some $P\in\mathcal{P}(X)$, then we do not have necessarily  $\PIV_{d'}^{\z}(X\to Y)=\PIEV_{d'}(X_P\to Y)$ for $d'\neq d$. 
For a counterexample, see Appendix  \ref{different}.
\end{Remark}

\subsection{PEACE, SPACE, and APACE}\label{PEACE, SPACE and APACE}
To calculate $\PIV_d^{\z}(X\to Y)$ (see Equation~\ref{piv}), we move along different partitions of $X$, and measure the highest interventional changes in $Y$ values by changing $X$ values while keeping $\Z=\z$. Now, if we only consider the slight changes of $X$ values, then we come up with $\PIEV_d^{\z}(X\to Y)$ (see Equation~\ref{piev}). Hence, the Probabilistic Easy vAriational Causal Effect (PEACE) of degree $d$ of $X$ on $Y$  is defined as
$\PEACE_d(X\to Y)=\E_{\Z}\left(\PIEV_d^{\z}(X\to Y)\right)$. 
We have the following observation:
\begin{Observation}
The following equalities hold:
\begin{align*}
	\PEACE_d(X\to Y) &= \sum_{i=1}^l\PEACE_d(X_{(x_{i-1},x_i)}\to Y),\\
	\PACE_d(X\to Y) &=\max\left\{\PEACE_d(X_P\to Y): P\in\mathcal{P}(X)\right\},\\
	\PACE_d(X\to Y) &=\max\left\{\sum_{i=1}^{n_P}\PEACE_d\left(X_{\left(x_{i-1}^{(P)},  x_i^{(P)}\right)}\to Y\right): P\in\mathcal{P}(X)\right\},
\end{align*}
where $X_{(x,x')}$ is the restriction of $X$ to the partition $\{x, x'\}$ for any two values $x$ and $x'$ of $X$ with $x<x'$. 
\end{Observation}
Let us consider some other patterns of changes  for $X$ values. First, assume that we only consider the single change from $X=x$ to $X=x'$ for $x<x'$. Then, we define the \textit{supremum probabilistic interventional variation} of degree $d$ of $Y$ with respect to $X$ keeping $\Z=\z$ as:
\begin{align*}
\SPIV_d^{\z}(X\to Y)&=\max_{\substack{x,x'\in\mathrm{Supp}(X)\\x<x'}}\PIEV_d^{\z}(X_{(x,x')}\to Y)\\
&= \max_{\substack{x,x'\in\mathrm{Supp}(X)}}\left|g_{in}(x,\z)-g_{in}(x',\z)\right|\P(x|\z)^d\P(x'|\z)^d.
\end{align*}
Consequently, we define the Supremum  Probabilistic vAriational Causal Effect (SPACE) of  degree  $d$ of $X$ on $Y$  as 
$\SPACE_d(X\to Y)=\E_{\Z}\left(\SPIV_d^{\z}(X\to Y)\right)$.
\begin{Observation}We have that
\[\SPACE_d(X\to Y)\ge \max_{\substack{x,x'\in\Supp(X)\\ x<x'}}\PEACE_d(X_{(x,x')}\to Y).\]
\end{Observation}
Next, consider that in our analysis, we are interested in examining every possible change from \(X = x\) to \(X = x'\) for any \(x, x' \in \mathrm{Supp}(X)\). This comprehensive approach ensures that the causal metric accounts for all possible interventional changes. Then, we define the \textit{aggregated probabilistic  interventional variation} of $Y$ with respect to $X$ keeping $\Z=\z$ as follows:
\begin{align*}
\APIV_d^{\z}(X\to Y)&=\sum_{\substack{x,x'\in\mathrm{Supp}(X)\\x<x'}}\PIEV_d^{\z}(X_{(x,x')}\to Y)\\
&= \sum_{\substack{x,x'\in\mathrm{Supp}(X)\\x<x'}}\left|g_{in}(x,\z)-g_{in}(x',\z)\right|\P(x|\z)^d\P(x'|\z)^d.
\end{align*}
Therefore, the Aggregated Probabilistic vAriational Causal Effect (APACE) of $X$ on $Y$  of degree $d$ is defined as 
$\APACE_d(X\to Y)=\E_{\Z}\left(\APIV_d^{\z}(X\to Y)\right)$.

\begin{Theorem}\label{mainforspaceapaceandpeace}
The following statements hold:
\begin{enumerate}
	\item SPACE and APACE satisfy all our postulates for defining a DCE, although PEACE satisfies all postulates but  \ref{restrction does not increase} (for a counterexample of \ref{restrction does not increase} , see Example S6.1).
	\item 
	If $X$ is binary, then 
	\begin{align*}
		\PACE_d(X\to Y)&= \PEACE_d(X\to Y)=\SPACE_d(X\to Y) = \APACE_d(X\to Y).
	\end{align*}
\end{enumerate}
\end{Theorem}
\begin{proof}
The first part could be shown similarly to the proof of Theorem \ref{checkingpostulatesforpace}. Also, the proof of the second part is straightforward. 
\end{proof}
We will refer to  $\PIV_d^{\z}(X\to Y), \PIEV_d^{\z}(X\to Y), \SPIV_d^{\z}(X\to Y)$ and $\APIV_d^{\z}(X\to Y)$ as  \uline{\textit{variations}} or \uline{\textit{variational formulas/metrics}}. Also, we refer to  $\PACE_d(X\to Y)$, $\PEACE_d(X\to Y)$, $\SPACE_d(X\to Y)$ and $\APACE_d(X\to Y)$ as \uline{\textit{variational DCEs}}. 

In the following theorem, we provide the relationship between the  different types of variational DCEs defined above.
\begin{Theorem}\label{discreteinequalities} The followings hold:
\begin{small}
	\begin{enumerate}
		\item 
		$\PEACE_d(X\to Y)\le\PACE_d(X\to Y)\le  \APACE_d(X\to Y)$,
		\item 
		$\SPACE_d(X\to Y)\le\PACE_d(X\to Y)\le  \APACE_d(X\to Y)$.
	\end{enumerate}
\end{small}
\end{Theorem}
\begin{proof}
See Appendix  \ref{3.10}.
\end{proof}
See Appendix  \ref{ex3.10} for a setting  that all inequalities in Theorem \ref{discreteinequalities} could be strict.  Besides, one could easily find many examples with $\PEACE_d(X\to Y)>\SPACE_d(X\to Y)$ and many example with $\PEACE_d(X\to Y)<\SPACE_d(X\to Y)$.

One could ask \uline{``Which of the aforementioned variational DCE metrics is more appropriate?''} it depends on what best suits our needs. For instance, if the highest interventional changes of $Y$ values under a single change from $X=x$ to $X=x'$ satisfies our need, we would better use SPACE. Otherwise, if we are somehow   interested in a weighted sum of interventional changes of $Y$ values under different changes of $X$, PACE, PEACE, and APACE are preferred. However, PEACE does not completely satisfy our postulates as we stated in Theorem \ref{mainforspaceapaceandpeace}.

Also, the generalization of \text{APACE} to the continuous case is subtle. We can define $\APIV_d^{\bm{z}}(X \to Y)$ in the continuous case through two distinct methodologies. Let us explore the first approach. Suppose that $X$, $\bm{Z}$, and $Y$ are continuous variables with $\operatorname{Supp}(X) \subseteq [a, b]$. We might naturally define
\begin{align*}
\APIV_d^{\bm{z}}(X \to Y) &:= \sup_{\substack{A \subseteq \operatorname{Supp}(X) \\ A \text{ is finite}}} \APIV_d^{\bm{z}}(X_A \to Y),\\
\APIV_d^{\bm{z}}(X_A\to Y)&=	\sum_{\substack{x,x'\in A,\\x< x'}}|g_{in}(x',\bm{z})-g_{in}(x,\bm{z})|f(x'|\bm{z})^df(x|\bm{z})^d,
\end{align*}
where $f(\,\cdot\,|\z)$ is the probability density function of $X$ given $\Z=\z$.

The following proposition demonstrates that with the first approach, $\APIV_d^{\bm{z}}(X \to Y)$ in the continuous case is either $0$ or $\infty$, making it unsuitable as a DCE metric for continuous random variables. 

\begin{Proposition}
Assume that $g_{in}(\cdot,\bm{z})$ is continuous, and there exist numbers $\alpha,\beta \in[a,b]$ with  $\left(g_{in}(\alpha,\bm{z})- g_{in}(\beta,\bm{z})\right)f(\alpha|\bm{z})f(\beta|\bm{z})\neq 0$. Then, by the above definition, we have that $\APIV_d^{\bm{z}}(X\to Y)=\infty$. 	
\end{Proposition}
\begin{proof}
See Appendix  \ref{infty}.
\end{proof}

In transitioning to the second approach, consider that if $X'$ is an independent copy of $X$ given $\Z$, then in the discrete case, it holds that 
\begin{align*}
\APIV_d^{\bm{z}}(X \to Y) &= \frac{1}{2} \mathbb{E}_{(x,x')\sim (\P_{X}^d, \P_{X}^d)} \bigl[ \left| g_{\text{in}}(x,\bm{z}) - g_{\text{in}}(x',\bm{z}) \right|\mid \z \bigr].
\end{align*}
We observe that $\P_X^d$ does not constitute a probability mass function, as the sum of the values governed by this function may not sum to 1. Nonetheless, we can still apply the concept of expected value to these types of functions.
This insight allows us to naturally generalize this metric to the continuous case just by using the definition of the expected value in the continuous setting:
\[
\APIV_d^{\bm{z}}(X \to Y) = \frac{1}{2} \int_{-\infty}^{\infty} \int_{-\infty}^{\infty} \left| g_{\text{in}}(x,\bm{z}) - g_{\text{in}}(x',\bm{z}) \right| f(x|\bm{z})^d f(x'|\bm{z})^d \, \mathrm{d}x \, \mathrm{d}x'.
\]

\begin{Remark}
For a generalized idea to define variational DCEs in the discrete case using matrices, see Section~S13.
\end{Remark}

\subsection{Positive and Negative Interventional Variational Causal Effects}
To measure the positive or the negative interventional changes of $Y$ while increasing the value of $X$ and keeping $\Z$ value constant, we define the positive and the negative versions of PACE, PEACE, SPACE, and APACE. To do so,  we define
\begin{align*}
\PIEV_d^{\z}(X_P\to Y)^{\bm{+}}&:=\sum_{i=1}^{n_P}\left(g_{in}(x_i^{(P)},\z)-g_{in}(x_{i-1}^{(P)},\z)\right)^{\bm{+}}\P(x_i^{(P)}|\bm{z})^d\P(x_{i-1}^{(P)}|\bm{z})^d, \\
\PIEV_d^{\bm{z}}(X\to Y)^{\bm{+}}&:=	\PIEV_d^{\z}(X_{\Supp(X)}\to Y)^{\bm{+}},
\end{align*}
where $P\in\mathcal{P}(X)$, and we assume that $\Supp(X)=\{x_0,\ldots, x_l\}$ in which $x_0<\cdots<x_l$. Other variational metrics and  negative versions are similarly defined.
In the following theorem, we provide the relationships between the positive, the negative, and the (absolute) variations.
\begin{Theorem}\label{signeddiscereteinequalities}
The followings hold:
\begin{align*}
	\PIEV_d^{\bm{z}}(X_P\to Y)&=\PIEV_d^{\bm{z}}(X_P\to Y)^{\bm{+}}+\PIEV_d^{\bm{z}}(X_P\to Y)^{\bm{-}},\quad P\in\mathcal{P}(X),\\
	\PIV_d^{\bm{z}}(X\to Y)&\le \PIV_d^{\bm{z}}(X\to  Y)^{\bm{+}}+\PIV_d^{\bm{z}}(X\to Y)^{\bm{-}},\\
	\PIV_d^{\bm{z}}(X\to Y)&\ge \max\left\{\PIV_d^{\bm{z}}(X\to  Y)^{\bm{+}},\PIV_d^{\z}(X\to Y)^{\bm{-}}\right\},\\
	\SPIV_d^{\bm{z}}(X\to Y)&=\max\left\{ \SPIV_d^{\bm{z}}(X\to  Y)^{\bm{+}},\SPIV_d^{\bm{z}}(X\to Y)^{\bm{-}}\right\},\\
	\APIV_d^{\bm{z}}(X\to Y)&=	\APIV_d^{\bm{z}}(X\to  Y)^{\bm{+}}+	\APIV_d^{\bm{z}}(X\to Y)^{\bm{-}}.
\end{align*}
\end{Theorem}
\begin{proof}
It is straightforwrad. 
\end{proof}

Now, we define 	$\PACE_d(X\to Y)^{\bm{+}} $ as follows:
\begin{small}\begin{align*}
	\PACE_d(X\to Y)^{\bm{+}}&:=\mathbb{E}_{\bm{Z}}\left(\PIV_d^{\bm{z}}(X\to Y)^{\bm{+}}\right)=\sum_{\bm{z}\in\Supp(\bm{Z})}\PIV_d^{\bm{z}}(X\to Y)^{\bm{+}}\P_{\bm{Z}}(\bm{z}).
	\end{align*}\end{small}
	The positive versions of other variational DCEs and also  negative versions are defined similarly.
	
	The following corollary of Theorem \ref{signeddiscereteinequalities} could be easily shown by taking the expected value with respect to $\bm{Z}$ from both sides of each item in Theorem \ref{signeddiscereteinequalities}.
	\begin{Corollary}\label{corsigneddiscreteinequalities}
The followings hold:
\begin{align*}
	\PACE_d(X\to Y)&\le 	\PACE_d(X\to Y)^{\bm{+}}+	\PACE_d(X\to Y)^{\bm{-}},\\
	\PACE_d(X\to Y)&\ge 	\max\{\PACE_d(X\to Y)^{\bm{+}},	\PACE_d(X\to Y)^{\bm{-}}\},\\
	\PEACE_d(X\to Y)&=\PEACE_d(X\to Y)^{\bm{+}}+\PEACE_d(X\to Y)^{\bm{-}},\\
	\SPACE_d(X\to Y)&\le\max\{\SPACE_d(X\to Y)^{\bm{+}},\SPACE_d(X\to Y)^{\bm{-}}\},\\
	\APACE_d(X\to Y)&=\APACE_d(X\to Y)^{\bm{+}}+\APACE_d(X\to Y)^{\bm{-}}.
\end{align*}
\end{Corollary}

The example in Appendix  \ref{ex3.14} shows that the inequalities in Theorem \ref{signeddiscereteinequalities} and Corollary~\ref{corsigneddiscreteinequalities} could be strict. 

One could show the following theorem similar to the proof of Theorem~\ref{checkingpostulatesforpace} and Theorem~\ref{mainforspaceapaceandpeace}.
\begin{Theorem}\label{nemidoonam}
Fixing $\bm{Z}=\bm{z}$,  the increase  in the value of $X$ (if possible) from the value $x$ to $x'$ does not causally increase (resp. decrease) the value of $Y$ for any $x,x'\in\Supp(X)$ with $x\le x'$, if and only if one of the followings holds:
\begin{align*}
	&\mathcal{P}\mathcal{V}_d^{\bm{z}}(X\to Y)^{\bm{\epsilon}}=0,& \mathcal{P}\mathcal{E}\mathcal{V}_d^{\bm{z}}(X\to Y)^{\bm{\epsilon}}=0,\\ &\mathcal{S}\mathcal{P}\mathcal{V}_d^{\bm{z}}(X\to Y)^{\bm{\epsilon}}=0,& \mathcal{A}\mathcal{P}\mathcal{V}_d^{\bm{z}}(X\to Y)^{\bm{\epsilon}}=0,
\end{align*}
where $\bm{\epsilon}=\bm{+}$ and $\bm{\epsilon}=\bm{-}$, when we talk about the increase and the decrease, respectively.
\end{Theorem}

\subsection{Natural Availability of Changing}

With a policy for modifying the values of the treatment variable at our disposal, we can calculate the corresponding changes in the outcome. In our methodology, for example, this policy could involve moving along a partition of the treatment values. By examining the changes in treatment values, we can assess how these changes manifest in reality, outside of our experimental setup. Thus, the concept of "natural availability of change" serves as a metric to measure this real-world variability.

For each of the variations, we define the corresponding natural availability of changing of degree $d$ of $X$ keeping $\Z=\z$, to be the variation of degree $d$ of $X$ with respect to itself keeping $\Z=\z$. More precisely, we define the \textit{probabilistic interventional easy natural availability of changing} ($\PIENAC$), \textit{probabilistic interventional  natural availability of changing} ($\PINAC$), \textit{supremum probabilistic interventional natural availability of changing }($\SPINAC$), and \textit{aggregated probabilistic interventional natural availability of changing} ($\APINAC$) of  degree $d$ of $X$ keeping $\Z=\z$ as follows:
\begin{align*}	
\PIENAC_d^{\z}(X_P)&:=\sum_{i=1}^{n_P}\left|x_i^{(P)}-x_{i-1}^{(P)}\right| \P(x_i^{(P)}|\z)^d\P(x_{i-1}^{(P)}|\z)^d,\quad P\in\mathcal{P}(X),\\
\PIENAC_d^{\z}(X)&:=\PIENAC_d^{\z}(X_{\Supp(X)}),\\
\PINAC_d^{\z}(X)&:=\max_{P\in\mathcal{P}(X)}\PIENAC_d^{\z}(X_P),
\\
\SPINAC_d^{\z}(X)&:=\max_{\substack{x,x'\in\Supp(X)\\ x<x'}}\PIENAC_d^{\z}(X_{(x,x')}),\\
\APINAC_d^{\z}(X)&:=\sum_{\substack{x,x'\in\Supp(X)\\ x<x'}}\PIENAC_d^{\z}(X_{(x,x')}).
\end{align*}
Finally, we define $\PACE_d(X|\Z),\PEACE_d(X|\Z), \SPACE_d(X|\Z)$ and $\APACE_d(X|\Z)$ as the expected values of the natural availability of changing defined above, respectively. The positive and the negative versions of these metrics are defined similarly. 

\subsection{Identifiability of our Variational DCE Metrics}\label{Identifiability of our Variational DCE Formulas}
Given an SEM and the joint probability distribution of observable and measurable random variables $X$ and $\Z$, we can calculate our variational DCEs. However, real-world problems often include unobserved random variables, potentially unmeasurable and might have \textit{proxies}\footnote{A proxy for an unobserved random variable $U$ is an observable variable or vector addressing $U$. For instance, if $U$ represents socio-economic status, a proxy could include components like annual salary, zip code, degree of education, and occupation.}. Therefore, let $Y = g(X, \Z, \U_Y)$, where $\U_Y$ is unobserved.
Roughly speaking, we say that a quantity defined by a formula $\mathcal{F}$, which is associated with an SEM, is \textbf{identifiable} under the assumption $\mathcal{A}$ if the value of $\mathcal{F}(S)$ can be uniquely determined using only the SEM itself and the observed variables, for any SEM \(S\) that satisfies assumption $\mathcal{A}$. This means that the result of $\mathcal{F}$ does not depend on any unobserved (latent) variables and is the same across all SEMs that align with $\mathcal{A}$ and the available data.

\begin{Definition}
We say that $g_{in}(X,\Z,\U_Y)$ is \textit{separable} with respect to $\Z$, whenever there exist two functions $g_{in}^{(1)}(X,\Z)$ and $g_{in}^{(2)}(\Z,\U_Y)$ in such a way that
$g_{in}(X,\Z,\U_Y)=g_{in}^{(1)}(X,\Z)+g_{in}^{(2)}(\Z,\U_Y)$.
\end{Definition}
\begin{Observation} Assume that $g_{in}(X,\Z,\U_Y)$ has the partial derivative with rspect to $X$. 
Then,  $\left(\partial g_{in}/\partial X\right)(X,\Z,\U_Y)$ does not depend on $\U_Y$ (i.e., it is a function of $X$ and $\Z$) if and only if $g_{in}(X,\Z,\U_Y)$ is separable with respect to $\Z$.
\end{Observation}
In the following theorem and its corollary, we provide an identifiability criterion for our variational DCEs while considering a partition of $\Supp(X)$ with two elements. These results can be naturally generalized to all of our variational DCEs.
\begin{Theorem}\label{identifiabilitymain}
Assume that 
\begin{enumerate}
	\item $g_{in}(X,\Z,\U_Y)$ is separable with respect to $\Z$, and
	\item $X$ and $\Z$ are independent given $\U_Y$ (in such a way that $\P(\z|\u_Y)\neq 0$ for any $(\Z,\U_Y)=(\z,\u_y)$).
\end{enumerate} 
Then, $\PEACE_d(X_{(x,x')}\to Y)$ is identifiable for any $x,x'\in\Supp(X)$ with $x<x'$. Further, assume that we have the following extra assumptions:
\begin{enumerate}
	\setcounter{enumi}{2}
	\item $Y_{x,\z}$ and $X$ are independent given $\Z$. 
	\item $Y_{x,\z}$ is a one-to-one function of $\left(\U_Y\right)_{x,\z}$, and
	\item $Y_{x,\z}$ and $\Z$ are independent.
\end{enumerate}
Then, we have that
\[\PEACE_d(X_{(x,x')}\to Y)=\E_{\Z}\biggl(\bigl|\E(Y|x',\z)-\E(Y|x,\z)\bigr|\P(x'|\z)^d\P(x|\z)^d\biggr).\]
\end{Theorem}
\begin{proof}
See Appendix  \ref{3.18}.
\end{proof}

\begin{Corollary}\label{cor3}
Assume that $g(X,\Z,\U_Y)$ is linear; namely $g(X,\Z,\U_Y)=\alpha X+ \Z\bm{\mathrm{\beta}} + \gamma U_Y$, where $\bm{\mathrm{\beta}}$  is a column vector, and $\gamma$ is a scalar with $\gamma\neq 0$. If 
\begin{itemize}
	\item  $X$ and $\Z$ are independent given $\U_Y$ (in such a way that $\P(\z|\u_Y)\neq 0$ for any $(\Z,\U_Y)=(\z,\u_y)$).
\end{itemize}
Then, $\PEACE_d(X_{(x,x')}\to Y)$ is identifiable for any $x,x'\in\Supp(X)$ with $x<x'$. Further, assume that we have the following additional assumptions:
\begin{enumerate}
	\item $Y_{x,\z}$ and $X$ are independent given $Z$, and
	\item $Y_{x,\z}$ and $\Z$ are independent. 
\end{enumerate}
Then,  we have that
\begin{align*}
	\PEACE_d(X_{(x,x')}\to Y)& =\alpha\PEACE_d(X|\Z)=\alpha (x'-x)\E_{\Z}\bigl(w_d(x',x,\z)\bigr),\\
	w_d(x',x,\z)& = \P(x'|\z)^d\P(x|\z)^d.
\end{align*}
\end{Corollary}

\section{The MEAN Versions of PEACE, PACE and APACE}\label{The MEAN Versions of PEACE, PACE and APACE}
Within the framework of the PEACE metric, we account for minor variations in the value of the treatment variable \(X\) along the sequence \(x_0, x_1, \ldots, x_l\), enabling the assessment of causal outcome changes. We calculate the natural availability of each incremental change from one treatment value to the next (e.g., \(x_0\) to \(x_1\)). The  direct total causal variations of \(X\) on \(Y\), with respect to these slight incremental changes, are then calculated as PEACE. However, the availability of these incremental changes is considered in a context where any type of change in the values of \(X\) is possible, leading to small availabilities for these increments. By allowing only incremental changes, we can calculate the new availabilities of changes for these increments. To do so, we define the MEAN PEACE,  $\overline{\PEACE}_d(X\to Y)$, as the mean of the following MEAN VARIATION:
\[\overline{\PIEV}_d^{\z}(X\to Y):=\sum_{i=1}^l|g_{in}(x_i,\z) -g_{in}(x_{i-1},\z)|\P_d(x_i,x_j\mid \z)\]
where
\[\P_d(x_i,x_j\mid \z) := \frac{\P(x_i\mid \z)^d\P(x_j\mid \z)^d}{\sum_{i=1}^l \P(x_i\mid \z)^d\P^d(x_{i-1}\mid \z)^d},\quad j=0,1,\ldots,l.\]
Similarly, we can define the MEAN PACE and the MEAN APACE as well as the positive and the negative versions of all of these MEAN metrics. 

It is important to note  that for a term $|g_{in}(x',\z)-g_{in}(x,\z)|\P(x'\mid \z)^d\P(x\mid \z)^d$ in our proposed DCE metrics, we call $|g_{in}(x',\z)-g_{in}(x,\z)|$ the causal part and $\P(x'\mid \z)^d\P(x\mid \z)^d$ its availability part of degree $d$.  The following points are notable concerning these mean metrics:
\begin{enumerate}
\item If $X$ is binary, then the mean versions of PACE, PEACE and APACE are all equal to 
$\E_{\Z}\bigl[\bigl|\ACDE(X\to Y\mid \Z)\bigr|\bigr]$.
\item Roughly, if in the  $\overline{\PIEV}_d^{\z}(X\to Y)$, the overall causal effect of the terms associated with more available changes exceeds that of the other terms, then MEAN PEACE is an increasing function of \(d\). Otherwise, it is decreasing.
\item In the $\overline{\PIEV}_d^{\z}(X\to Y)$, as \(d \to \infty\), then $\overline{\PIEV}_d^{\z}(X\to Y)$ tends to the causal effect associated with the most available change. If the most available change is not unique, this limit is equal to their average.
\item Similarly, in the $\overline{\PIEV}_d^{\z}(X\to Y)$, as \(d \to -\infty\), then $\overline{\PIEV}_d^{\z}(X\to Y)$ tends to the causal effect associated with the least available change. If the least available change is not unique, this limit is equal to their average.
\end{enumerate}
\begin{Example}
Assume that $\Supp(X) = \{x_0, x_1, x_2\}$ and for $\Z = \z$, we define $a = \mathbb{P}(x_0 \mid \z) \mathbb{P}(x_1 \mid \z)$ and $b = \mathbb{P}(x_1 \mid \z) \mathbb{P}(x_2 \mid \z)$, and
\begin{align*}
	|g_{\text{in}}(x_1, \z) - g_{\text{in}}(x_0, \z)| = \alpha, \quad |g_{\text{in}}(x_2, \z) - g_{\text{in}}(x_1, \z)| = \beta.
\end{align*}
Then, the function $f(d) = \overline{\PIEV}_d^{\z}(X \to Y) = (\alpha a^d + \beta b^d)/(a^d + b^d)$. We note that
\[
f'(d) = \frac{a^d b^d (\alpha - \beta) \log\left(\frac{a}{b}\right)}{(a^d + b^d)^2}.
\]
Now, assume that the change of degree $d$ from $x_0$ to $x_1$ is more available than that from $x_1$ to $x_2$. This implies $a > b$. If the causal effect corresponding to this more available change, $\alpha$, is also greater than the other one, $\beta$, then $f'(d) > 0$, and hence $f$ is increasing. Otherwise, if the causal effect corresponding to the less available change is higher, then $f'(d) < 0$, and hence $f(d)$ is decreasing.
\end{Example}

\section{Simpson's paradox}\label{simpson}
Simpson’s paradox is a significant issue in causal reasoning. In some real-world problems, a trend might be observed in several subgroups of a dataset but disappear or reverse when considering the entire dataset. This paradox can occur when different groups have varying sample sizes and/or different confounding variables. The mathematical justification for this paradox is based on the simultaneous occurrence of the following inequalities:
\[
1 \ge \frac{a_i}{b_i} > \frac{c_i}{d_i}, \quad i=1,\ldots,n, \quad \frac{a_1+\cdots+a_n}{b_1+\cdots+b_n} < \frac{c_1+\cdots+c_n}{d_1+\cdots+d_n}.
\]
To illustrate this, consider the example shown in Table~\ref{table}, which examines the gender effect on admission rates among applicants to the University of California, Berkeley, for the autumn of 1973. Assume that the department acts as a confounder for both the admission rate and the gender of the applicants. For simplicity, we denote "man" and "woman" by 0 and 1, respectively. Additionally, we use the random variables $X$, $Z$, and $Y$ to represent gender, department, and admission rate, respectively. Assume that the identifiability criteria given in Theorem \ref{identifiabilitymain} are satisfied. 
One could see that:
\begin{align*}
\PACE_d(X\to Y)^{\bm{+}}&=0.2\left(\frac{933}{4526}\right)\left(\frac{89100}{870489}\right)^d+0.05\left(\frac{585}{4526}\right)\left(\frac{14000}{342225}\right)^d\\
&+0.02\left(\frac{792}{4526}\right)\left(\frac{156375}{627264}\right)^d+0.01\left(\frac{714}{4526}\right)\left(\frac{127193}{509796}\right)^d,\\
\PACE_d(X\to Y)^{\bm{-}}&=0.03\left(\frac{918}{4526}\right)\left(\frac{192725}{842724}\right)^d+ 0.04\left(\frac{584}{4526}\right)\left(\frac{75063}{341056}\right)^d.
\end{align*}
As shown in Figure \ref{simpsonfig}, the direct causal effect of being admitted as a woman is greater than that of being a man. Thus, our methodology yields an intuitively consistent result for this example, resolving the paradox.
\begin{table}
\centering
\caption{The data for the concrete example provided in Section \ref{simpson}.}\label{table}
\begin{tabular} { | c | c | c | c | c | c| c |}
	\hline
	\multirow{2}{*}{\footnotesize{Department} } 	& \multicolumn{2} { | c | }{\footnotesize{All}}& \multicolumn{2} { | c | }{\footnotesize{Men}}& \multicolumn{2} { | c | }{\footnotesize{Women}}\\
	\cline{2-7}
	& \scriptsize{Applicants} & \scriptsize{Admitted}&\scriptsize{Applicants} & \scriptsize{Admitted}&\scriptsize{Applicants} & \scriptsize{Admitted}\\
	\hline
	A&933&64\%& 825&62\%&108&\textcolor{red}{82\%}\\
	\hline
	B&	585&	63\%&	560	&63\%&	25&	\textcolor{red}{68\%}\\
	\hline
	C&	918&	35\%&	325	&\textcolor{red}{37\%}&	593	&34\%\\
	\hline
	D&	792&	34\%&	417	&33\%&	375&	\textcolor{red}{35\%}\\
	\hline
	E&	584&	25\%&	191	&\textcolor{red}{28\%}	&393&	24\%\\
	\hline
	F&	714	&6\%&	373	&6\%&	341	&\textcolor{red}{7\%}\\
	\hline
	Total&	4526&	39\%&	2691&	\textcolor{blue}{45\%}&	1835&	30\%\\
	\hline
\end{tabular}
\end{table}
\pgfplotsset{compat = newest}
\begin{figure}
\caption{\raggedright \small These plots indicate that, in the example given in Section \ref{simpson}, from a causal perspective, the admission rate for women is higher regardless of whether the degree \(d\) is high or low.
}\label{simpsonfig}
\subfloat[\raggedright  \scriptsize The red and the blue graphs show that $\PACE_d(X\to Y)^{\bm{+}}$ and $\PACE_d(X\to Y)^{\bm{-}}$.]{
	\begin{tikzpicture}
		\begin{axis}[ scale =0.8,
			xmin = 0, xmax = 2,
			ymin = 0, ymax = 0.06,
			width=0.5\textwidth, 
			height=0.375\textwidth, 
			xlabel = $d$, ylabel = $\PACE_d(X\to Y)^{\bm{\epsilon}}$
			]
			\addplot[
			domain = 0:2,   samples = 200,
			smooth,
			thick,
			blue,
			] {(0.2*(933/4526)*((89100/870489)^x)+0.05*(585/4526)*((14000/342225)^x)+0.02*(792/4526)*((156375/627264)^x)+0.01*(714/4526)*((127193/509796)^x))};
			\addplot[
			domain = 0:2,   samples = 200,
			smooth,
			thick,
			red,
			] {(0.03*(918/4526)*((192725/842724)^x)+ 0.04*(584/4526)*((75063/341056)^x))};
			\legend{
				$\bm{\epsilon} = \bm{+}$,
				$\bm{\epsilon} = \bm{-}$
			}
		\end{axis}
\end{tikzpicture}}
\subfloat[\raggedright \scriptsize The natural availabilities of changing functions decrease at the rate visible in the plot. This implies a corresponding decrease in the PEACE graphs.
]{
	\begin{tikzpicture}
		\begin{axis}[scale =0.8,
			xmin = 0, xmax = 2,
			ymin = 0, ymax = 0.6,
			width=0.5\textwidth, 
			height=0.375\textwidth, 
			xlabel = $d$, ylabel = $\PACE_d(X\mid \Z )^{\bm{\epsilon}}$
			]
			\addplot[
			domain = 0:2,   samples = 200,
			smooth,
			thick,
			blue,
			] {((933/4526)*((89100/870489)^x)+(585/4526)*((14000/342225)^x)+(792/4526)*((156375/627264)^x)+(714/4526)*((127193/509796)^x))};
			\addplot[
			domain = 0:2,   samples = 200,
			smooth,
			thick,
			red,
			] {((918/4526)*((192725/842724)^x)+ (584/4526)*((75063/341056)^x))};
			\legend{
				$\bm{\epsilon} = \bm{+}$,
				$\bm{\epsilon} = \bm{-}$
			}
		\end{axis}
	\end{tikzpicture}
}
\end{figure}

\section{Comparing our Methodology to the Pearl Framework}\label{Comparing our Framework to the Pearl Framework}

In our approach as well as the Pearl framework, the concept of intervention plays an important role to calculate the causal effect of a random variable $X$ on $Y$. However, compared to the Pearl framework, the concept of natural availability of changing  $X$ values, keeping $\Z$ constant, provides a distinguishable characteristic for our methodology.  We believe that in each subpopulation determined by $\Z$, the frequencies of applying/observing each of $X=x$ and $X=x'$ ($\P(x|\z)$ and $\P(x'|\z)$) affect what we expect to be the causal effect of $X$ on $Y$. \uline{As an example, we might expect that the rarity of some values  of $X$ has a small share on the causal effect of $X$ on $Y$ in some specific applications such as the systems that are robust with little noise.}
Nevertheless, in usual causal effect metrics used in the Pearl framework, after  statistical identifiability, only the frequencies of the aforementioned subpopulations affect the causal effect of $X$ on $Y$ as weights (i.e., in these metrics $\P(y|x,\z)$ and $\P(\z)$ are important, not $\P(x|\z)$). Our variational DCEs are relative to the types of changes we consider. In fact, to measure the variational DCE  of $X$ on $Y$, we may change the value of $X$ along a partition of $X$ (e.g., PEACE and PACE). Also, we may  just look for the maximum/sum of interventional changes of $Y$ under  single changes of $X$ (i.e., SPACE/APACE). In graphical causality by Pearl, a given DAG could lose some of its arrows when  intervention on a node is applied. In our approach in addition to the previous property, to measure the variational DCE of $X$ on $Y$ along a path, some mediator nodes could be replaced with their parents. However, the edges in the new graph might come from different functions (see Appendix  \ref{indirect}).

Our variational DCEs are comparable to the direct causal effect metrics used in the Pearl framework such as the average control direct effect (ACDE). If after the intervention on the value of $(X,\Z)$, $\U_Y$ remains unchanged, then we have that
\[g_{in}(x',\z,\u_Y)-g_{in}(x,\z,\u_Y)=\E(Y_{x',\z})-\E(Y_{x,\z}),\]
which is the ACDE of $X_{(x,x')}$ on $Y$. Hence, our variational DCEs and their mean versions seem to be derived from the weighted ACDEs. This is true, but \uline{these weights are not ordinary weights as they are natural availability of changing  $X$ values}. 

In the following observation, we provide  a relationship between ACDE and PACE when $d=0$ and $X$ is binary.
\begin{Observation}
If $X$ is binary, and after the intervention on the value of $(X,\Z)$, $\U_Y$ remains unchanged, then for any $d$, we have that 
\[\PACE_0(X\to Y)=\overline{\PACE}_d(X\to Y)=\E_{\Z}\bigl(|\ACDE\bigl((X\to Y)|\Z=\z\bigr)|\bigr).\]
\end{Observation}
We recall that when \(X\) is binary, all of our variational DCEs coincide. Furthermore, the mean versions of our variational DCEs also coincide.

Our point of view could be used to define some  new causal total  effect metrics such as the one in the following for each $P\in\mathcal{P}(X)$:
\begin{align}	
\stackrel{ACE}{\PEACE_d}(X_P\to Y)&:=\sum_{i=1}^{n_P}\left|\E(Y_{x_{i}^{(P)}})-\E(Y_{x_{i-1}^{(P)}})\right| \P(x_i^{(P)})^d\P(x_{i-1}^{(P)})^d\label{totaleffect}
\end{align}
The positive and the negative versions as well as the  mean versions of these causal metrics could be defined naturally as before. Moreover, the above idea could be used for other causal effect metrics as well (not only ACE).

\section{Investigating Some Examples}\label{Investigating Some Examples}
In Section \ref{simpson}, we explored Simpson's paradox as an example with a non-binary treatment. In this section, we investigate three examples with binary treatments using the methods of Pearl, Janzing et al., mutual information, conditional mutual information, and our own methodologies.

\subsection{A Binary Symmetric Channel}\label{BSC}

Consider a communication channel with a binary input $X$, a binary output $Y$, and noise $Z$ affecting $X$, such that $Y = g(X,Z) = X \oplus Z$, where $\oplus$ denotes the XOR operation. Assume that $X$ and $Z$ are independent and uniformly distributed. We now analyze the causal effect of $X$ on $Y$ using various metrics:

\begin{itemize}
\item \textbf{PACE:} 
\[
\PACE_d(X\to Y) = \mathbb{E}_Z(\PIV_d^z(X\to Y)) = \frac{1}{4^d}.
\]
This represents the maximum PACE among all possible distributions for $X$, as the maximum natural availability of changing for a binary treatment is achieved when the treatment is uniformly distributed.

\item \textbf{MEAN PACE:} 
\[
\overline{\PACE}_d(X\to Y) = \mathbb{E}_Z\left[\left|\ACDE\left((X\to Y)\mid Z=z\right)\right|\right] = 1.
\]
Thus, the MEAN PACE successfully captures a direct causal effect of $X$ on $Y$. In the binary case, the MEAN PACE does not depend on $d$, as each natural availability of changing the treatment value is considered among other availabilities. In this case, there is only one natural availability of change, which is eliminated when taking the average.

\item \textbf{ACDE:} 
\[
\ACDE\left((X\to Y)\mid Z\right) = 0.
\]
Clearly, the value of $Y$ changes when we fix the value of $Z$ and change the value of $X$, which confirms that $X$ is indeed a true cause of $Y$. Therefore, ACDE fails to capture a causal effect of $X$ on $Y$.

\item \textbf{CACE \& ACE:} We observe that the CACE of $X$ on $Y$ given $Z=z$ does not vanish for any $z \in \{0,1\}$, although the average CACE with respect to $Z$ (i.e., 
$\ACE(X \to Y)$) 
vanishes. Thus, while ACE fails to capture a causal effect of $X$ on $Y$, CACE can successfully capture it.

\item \textbf{KL Divergence Metric in Janzing et al. Model:} 
\[
\mathfrak{C}_{X\to Y} = D_{\kl}(\mathbb{P} \parallel \widetilde{\mathbb{P}}) = 1,
\]
where $D_{\kl}$ denotes the Kullback-Leibler divergence. This correctly quantifies a causal effect in this scenario.

\item \textbf{Mutual Information:} We have
$
\I(X; Y) = 0$, 
which indicates that the mutual information framework fails to capture a direct causal effect of $X$ on $Y$.

\item \textbf{Conditional Mutual Information:} We have that
$
\I(X; Y \mid Z) = 1$.
This correctly reflects a direct causal influence of $X$ on $Y$ when conditioned on $Z$.
\end{itemize}

\subsection{Investigating the Effect of a Rare Disease On Blood Pressure}\label{Investigating the Effect of a Rare Disease On Blood Pressure}

Assume that a rare disease causes a notable increase in blood pressure. Let us define the variables as follows:
\begin{align*}
X &= \begin{cases}
	1, & \text{presence of the disease}\\
	0, & \text{otherwise}
\end{cases},\quad
Y &= \begin{cases}
	1, & \text{blood pressure $>$ a given threshold}\\
	0, & \text{otherwise}
\end{cases}.
\end{align*}
There is a one-to-one correspondence between $X$ and $Y$, namely $Y = 1$ if and only if $X = 1$, implying that $Y = g(X) = X$. We now analyze the causal effect of $X$ on $Y$ using different metrics:
\begin{itemize}
\item \textbf{PACE:} 
\[
\PACE_d(X \to Y) = (p(1-p))^d, \quad p = \P(X = 1) > 0.
\]
Here, we observe that the PACE of degree $0$ is $1$ and the PACE of degree $1$ is $p(1-p)$. Since $p$ is very small, the natural availability of changing the value of $X$ from $0$ to $1$ is also small.  Given that the rare disease significantly increases blood pressure, higher values of $d$ might mislead us by yielding low values for the causal effect of $X$ on $Y$. Therefore, we should focus on lower values of $d$ to avoid underestimating the effect.

\item \textbf{MEAN PACE \& ACDE:} 
\[
\overline{\PACE}_d(X \to Y) = \ACDE(X \to Y) = 1.
\]
This shows that MEAN PACE and ACDE can capture a causal relationship between $X$ and $Y$. However, in observational studies involving rare cases, large sample sizes may be necessary to compute ACDE accurately. Furthermore, if we had a similar scenario where the rare case was not important, MEAN PACE and ACDE would still both equal 1, which would fail to differentiate between the two scenarios. 

We note that in this example, ACDE, ACE and CACE coincide. 

\item \textbf{Information Theoretic Causal Strength:} In all three information-theoretic methodologies, the causal strength of $X \to Y$ is given by the mutual information:
\[
\I(X; Y) = \h(p) = p \log \left( \frac{1}{p} \right) + (1-p) \log \left( \frac{1}{1-p} \right),
\]
which is very small due to the small value of $p$. Therefore, all three information-theoretic frameworks fail to accurately measure a direct causal effect of $X$ on $Y$ in this example, as they underestimate the importance of the rare events.
\end{itemize}

\subsection{Investigating the Causal Effects of Rain and Sprinkler on Wet Grass}\label{sprinkler}

Let $C$, $R$, $S$, and $W$ be binary random variables denoting two different opposite states for clouds, rain, the sprinkler, and the wetness of grass, respectively. More precisely, we have that:
\begin{align*}
&C=\left\{\begin{array}{ll}1,& \text{cloudy weather}\\
	0,& \text{otherwise}\end{array}\right.,&R=\left\{\begin{array}{ll}1,& \text{rain}\\
	0,& \text{otherwise}\end{array}\right.\\
&S=\left\{\begin{array}{ll}1,& \text{the sprinkler on}\\
	0,& \text{otherwise}\end{array}\right.,&W=\left\{\begin{array}{ll}1,& \text{wet grass}\\
	0,& \text{otherwise}\end{array}\right.\\
	\end{align*}
	We consider the additional assumption that $R$ and $S$ are independent given $C$. Assume that we have an SEM consisting of the following three equations, along with an additional equation describing \(V_3\) in terms of \(R\), \(S\), and some other unobserved factors:
	\begin{align*}
&R=f(C, V_1)=C\oplus V_1,\qquad
S=g(C, V_2)=C\oplus V_2,\\ &W=h(R,S,V_3)=\left\{\begin{array}{ll}1,&R=S=1,\\ R\oplus S\oplus V_3,& \text{otherwise}
\end{array}\right.
\end{align*}
Here, $V_1$, $V_2$, and $V_3$ could be binary random variables associated with rain clouds, seasons, and a mixture of positive and negative factors influencing the wetness of grass, respectively ($V_3$ is directly caused by $R$, $S$, and an unknown variable).  Assume that the probabilistic relationships between $C$, $R$, $S$ and $W$ are given in the following tables:
\[\begin{small}\begin{array}{ll}
	\begin{tabular}{cc}
		$\P(C=1)$ & $\P(C=0)$\\
		\hline
		0.5 & 0.5
	\end{tabular}, & \begin{tabular}{c|cc}
		$C$ & $\P(R=1\mid C)$ & $\P(R=0\mid C)$\\
		\hline
		1 & 0.8 & 0.2\\
		0 & 0.2 & 0.8
	\end{tabular}\\
	\begin{tabular}{c|cc}
		$C$ &$ \P(S=1\mid C)$ & $\P(S=0\mid C)$\\
		\hline
		1 & 0.1 & 0.9\\
		0 & 0.5 & 0.5
	\end{tabular}, &\begin{tabular}{c|ccc}
		$S$ & $R$& $\P(W=1\mid S,R)$ & $\P(W=0\mid S,R)$\\
		\hline
		1 & 1& 1 & 0\\
		1 & 0 & 0.9 & 0.1\\
		0 & 1 & 0.9 & 0.1\\
		0 & 0 & 0.01 & 0.99
	\end{tabular}
	\end{array}\end{small}\]
	The graphical view of this example is shown in Figure \ref{graph}.
	
	\begin{figure}
\begin{center}
	\caption{\raggedright \small{The graphical view of the relationships between the random variables in the example given in Section \ref{sprinkler}}}\label{graph}
	\vspace*{-2cm}
	\definecolor{ffqqqq}{rgb}{1,0,0}
	\definecolor{qqqqff}{rgb}{0,0,1}
	\definecolor{test}{rgb}{0.3,0.5,0.3}
	\begin{tikzpicture}[line cap=round,line join=round,>=triangle 45,x=1cm,y=1cm,scale=0.33]
		\clip(-23,-15) rectangle (32.4227984915779,9.979103805073944);
		\draw [rotate around={0:(-2,0)},line width=0.8pt,color=qqqqff] (-2,0) ellipse (2.826430586355788cm and 1.997175470379987cm);
		\draw [rotate around={0:(-8,-6)},line width=0.8pt,color=qqqqff] (-8,-6) ellipse (2.8264305863557855cm and 1.997175470379985cm);
		\draw [rotate around={0:(4,-6)},line width=0.8pt,color=qqqqff] (4,-6) ellipse (2.826430586355783cm and 1.9971754703799833cm);
		\draw [rotate around={0:(-2,-12)},line width=0.8pt,color=qqqqff] (-2,-12) ellipse (2.826430586355778cm and 1.9971754703799798cm);
		\draw [->,line width=0.8pt,color=qqqqff] (-3.9711346287901463,-1.431352340024542) -- (-8,-4.002824529620014);
		\draw [->,line width=0.8pt,color=qqqqff] (0.01733534146906468,-1.3988440423222885) -- (4,-4.002824529620016);
		\draw [->,line width=0.8pt,color=qqqqff] (-8,-7.997175470379984) -- (-3.6047275789027484,-10.355933626778373);
		\draw [->,line width=0.8pt,color=qqqqff] (4,-7.997175470379982) -- (-0.4263349821009008,-10.341009649955911);
		\draw [rotate around={0:(6,0)},line width=0.8pt,dash pattern=on 1pt off 1pt,color=ffqqqq] (6,0) ellipse (2.82643058635579cm and 1.9971754703799884cm);
		\draw [->,line width=0.8pt,dash pattern=on 1pt off 1pt,color=ffqqqq] (6,-1.9971754703799884) -- (5.6560094511397905,-4.381524158295875);
		\draw [rotate around={0:(-10,0)},line width=0.8pt,dash pattern=on 1pt off 1pt,color=ffqqqq] (-10,0) ellipse (2.8264305863557975cm and 1.9971754703799938cm);
		\draw [->,line width=0.8pt,dash pattern=on 1pt off 1pt,color=ffqqqq] (-10,-1.9971754703799938) -- (-9.660787293744807,-4.383970395736615);
		\draw [rotate around={0:(-2,-6)},line width=0.8pt,dash pattern=on 1pt off 1pt,color=ffqqqq] (-2,-6) ellipse (1.7cm and 1.8cm);
		\draw [->,line width=0.8pt,dash pattern=on 1pt off 1pt,color=ffqqqq] (-2,-7.997175470379982) -- (-2,-10);
		\draw (-4.2,0.85) node[anchor=north west] {Cloudy};
		
		\draw (5.3,0.75) node[anchor=north west] {$V_1$};
		\draw [->,line width=0.8pt,dash pattern=on 1pt off 1pt,color=ffqqqq] (-7.1,0) -- (-4.85,0);
		\draw (-11,.75) node[anchor=north west] {$V_2$};
		\draw [<-,line width=0.8pt,dash pattern=on 1pt off 1pt,color=ffqqqq] (-3.7,-6)--(-5.15,-6) ;
		\draw (-3,-5.1) node[anchor=north west] {$V_3$};
		\draw (-10.85,-5) node[anchor=north west] {Sprinkler};
		\draw [<-,line width=0.8pt,  dash pattern=on 1pt off 1pt,color=ffqqqq] (-0.3,-6)--(1.15,-6) ;
		\draw (2.4,-5.1) node[anchor=north west] {Rain};
		\draw (-4.5,-10.2) node[anchor=north west] {$\begin{array}{c}\text{Wet}\\ \text{Grass}\end{array}$};
	\end{tikzpicture}
\end{center}
\end{figure}
Assume that $\color{blue}{\P(V_3|R=1, S=1)\sim \B(p)}$, where $p\in[0,1]$. We now investigate the causal effects of $R$ and $S$ on $W$ using different metrics:
\begin{itemize}
\item \textbf{PACE:}
\begin{align*}
	\PACE_d(R\to W)&=0.6561\left(\frac{1307.9}{5314.41}\right)^d+
	0.0439\left(\frac{1.189}{19.2721}\right)^d\\
	&+0.3(0.07+0.3p)\left(\frac{0.021p}{(0.07+0.3p)^2}\right)^d,\\
	\PACE_d(S\to W)&=0.4761\left(\frac{66990}{279841}\right)^d+0.0239\left(\frac{24360}{228484}\right)^d\\
	&+0.5(0.082+0.18p)\left(\frac{0.0147p}{(0.082+0.18p)^2}\right)^d.
\end{align*}
\pgfplotsset{compat=1.16}

\begin{figure}
	\centering
	\subfloat[\raggedright \scriptsize Graph of $\PACE_d(S\to W)$ in red and $\PACE_d(R\to W)$ in blue as functions of $d$ and $p$, where $p=\P(V_3\mid R=1,S=1)$.\label{fig2}] {
		\begin{tikzpicture}[scale=1.2]
			\begin{axis}[
				xlabel=$d$,
				ylabel={$p$},
				zlabel style={rotate=-90},
				view={40}{30},
				xmin=0, xmax=1,
				ymin=0, ymax=1,
				zmin=0, zmax=0.85,
				small,
				]
				\addplot3[
				surf,
				domain=0:1,
				y domain=0:1,
				color=red, 
				opacity=0.7
				] 
				{0.4761*((66990/279841)^x) + 0.0239*((24360/228484)^x) + 0.5*(0.082+0.18*y)*((0.0147*y)/(0.082+0.18*y)^2)^x};
				
				\addplot3[
				surf,
				domain=0:1,
				y domain=0:1,
				color=blue, 
				opacity=0.7
				] 
				{0.6561*((1307.9/5314.41)^x) + 0.0439*((1.189/19.2721)^x) + 0.3*(0.07+0.3*y)*((0.021*y)/(0.07+0.3*y)^2)^x};
			\end{axis}
	\end{tikzpicture}}
	\subfloat[\raggedright \scriptsize The graphs of $\overline{\PACE}_d(S\to W)$  and $\overline{\PACE}_d(R\to W)$ as functions of $p$ are shown in red and blue, respectively.\label{figg2}]
	{
		\begin{tikzpicture}[scale =0.75]
			
			\begin{axis}[
				xmin = 0, xmax = 1,
				ymin = 0, ymax = 1,
				xlabel = $p$
				]
				\addplot[
				domain = 0:1,   samples = 200,
				smooth,
				thick,
				blue,
				] {0.702+0.09*x};
				\addplot[
				domain = 0:1,   samples = 200,
				smooth,
				thick,
				red,
				] {0.541+0.09*x};
				\legend{
					$R$,
					$S$
				}
			\end{axis}
		\end{tikzpicture}
	}
\end{figure}

Figure \ref{fig2} represents the PACEs of $R$ and $S$ on $W$ as the blue and red surfaces, respectively, with parameters $p$ and $d$. This figure  shows that 
$\PACE_d(R\to W) > \PACE_d(S\to W)$ 
for any fixed values of $p, d \in [0, 1]$. It is worth mentioning that for a fixed $0 \leq d \leq 1$, the PACEs of $R$ and $S$ on $W$ are both increasing functions with respect to $p$. However, beyond a certain degree $d$ greater than $1$, the behavior of the functions changes from being increasing to be decreasing as $p$ varies. The existance of $p$ in the obtained PACEs is due to the fact that we fix $V_3$ constant, when we want to calculate the direct causal effect of either $R$ or $S$ on $W$. When the PACE is an increasing function of $p$ for a fixed \(d\), it indicates that the higher the availability of \(V_3 = 1\) given $(R,S)=(1,1)$ is, the higher the PACE. Conversely, a decreasing function suggests that the less available \(V_3 = 1\) given $(R,S)=(1,1)$ is, the higher the PACE. If we examine the PACEs of \(R\) and \(S\) on \(W\) more closely, we observe that \(p\) is included in the third term. This term arises when we fix \((S, V_3) = (1, 1)\) and the treatment is \(R\), or when we fix \((R, V_3) = (1, 1)\) and the treatment is \(S\). Considering the case where \(R\) is the treatment (the analysis for \(S\) as the treatment follows similarly), the probability value \(p\) reflects the availability of changing the value of  \(R\) from \(R=0\) to \(R=1\) while keeping \((S, V_3)=(1,1)\). For a fixed big enough $d$, as \(p\) increases, the likelihood of \(R=1\) increases while the likelihood of \(R=0\) decreases. This shift makes the transition from \(R=0\) to \(R=1\) less available, thereby rendering \(\PACE_d(R\to W)\) a decreasing function of \(p\) for a fixed \(d\).

\item 
\textbf{MEAN PACE:}
\begin{align*}
	\overline{\PACE}_d(R\to W)&=0.6561+0.0439+0.3(0.07+0.3p) = 0.7021+0.09p,\\
	\overline{\PACE}_d(S\to W)&=0.4761+ 0.0239+0.041+0.09p=0.541+0.09p.
\end{align*}
Figure \ref{figg2} represents the MEAN PACEs, showing  $\overline{\PACE}_d(R\to W) > \overline{\PACE}_d(S\to W)$. As we see, both values are increasing as $p$ increases. This suggests that the likelihood of \(V_3=1\) when \(R=1\) and \(S=1\), in the absence of natural availabilities, increasingly contributes to the direct causal impact of \(R\) and \(S\) on \(W\).

\item 
\textbf{ACE \& ACDE:} Here we need to assume that the local Markov assumption is satisfied. We have that 
\begin{align*}
	\ACE(R\to W)&=	\ACDE\left((R\to W)|S\right)=\ACDE\left((R\to W)|S,V_3\right)=0.653,
\end{align*}
which is greater than
\begin{align*}
	\ACE(S\to W)&=	\ACDE\left((S\to W)|R\right)=\ACDE\left((S\to W)|R,V_3\right)=0.495.
\end{align*}
\item 
\textbf{Information Theoretic Causal Strength:}
Let $\widetilde{\P}$ and $\widehat{\P}$ be the post-cutting distributions of the DAG in Figure \ref{graph} after cutting the arrows $R\to W$ and $S\to W$, respectively. Then, 
the causal strengths ‌of $R\to W$ and $S\to W$ using Janzing et al. framework are as follows:
\begin{align*}
	\mathfrak{C}_{R\to W} = D_{\mathrm{KL}}(\P||\widetilde{\P})\approx 0.351431,\qquad \mathfrak{C}_{S\to W}=D_{\mathrm{KL}}(\P||\widehat{\P})\approx0.270828.
\end{align*}
Consequently, Janzing et al. framework as well as  PACE and ACDE calculates a higher direct causal effect for $R$ on $W$. Moreover, one could see that in the mutual information and the conditional mutual information frameworks, we have that
\[\begin{array}{ll}
	\I(R;W)\approx 0.2483275,
	&
	\I(S;W)\approx 0.125463,\\
	\I(R;W|S)\approx 0.49359151,
	&
	\I(S;W|R)\approx 0.37072701.
\end{array}\]
\end{itemize}
We observe that all of the causal metrics used provide a higher causal impact of \(R\) on \(W\) compared to \(S\) on \(W\). However, the difference between these two causal impacts is a bit larger when using our metrics. Since PACE depends on the availabilities of the treatment values, \(p = \P(V_3 = 1 \mid R = 1, S = 1)\) appears in the value of PACE. Moreover, both \(\P(V_3 = 1, S = 1)\) and \(\P(V_3 = 1, R = 1)\) are functions of \(p\), which implies that \(p\) also appears in the MEAN PACEs of \(R\) and \(S\) on \(W\).

\section{Generalized Methodologies}\label{general}
In this section, we introduce several generalizations of our variational DCEs from different perspectives.
\begin{itemize}
\item In our variational DCE metrics, we account for the rarity and frequency of the treatment values. However, there are situations where a very small portion of the population can significantly dominate well-known causal metrics such as ACE. To control the importance of the rarity or frequency of subpopulations, we introduce a new degree \(r\), in addition to the previously considered degree \(d\). Specifically, we replace the probability values \(\P(\z_0), \ldots, \P(\z_k)\) of observing all possible values of \(\Z\) with \(\P(\z_0)^r/A, \ldots, \P(\z_k)^r/A\), where 
\[
A = \P(\z_0)^r+ \cdots + \P(\z_k)^r.
\]
Thus, we define the generalized PACE of degree \((d, r)\) as follows:
\[
\PACE_d^{(r)}(X \to Y) := \mathbb{E}_{\z \sim \P_{\Z}^r/A} \left[\PIV_d^{\z}(X \to Y)\right].
\]
Similarly, we can define the generalized versions of other variational DCEs, including their mean, positive, and negative versions. By increasing the value of \(r\), we place more emphasis on the more frequent subpopulations determined by $\Z$. Conversely, lower values of \(r\) closer to 0 assign roughly equal importance to all subpopulations.
\item 
To generalize the natural availability of changing treatment values, we can consider a random process \(X^{\z}\) defined over two time units: one before making the change in the value of \(X\), denoted by \(X^{\z}_0\), and the other after the change, denoted by \(X^{\z}_1\). In our variational DCEs, we assumed that the probability of observing \(X^{\z}_0 = x\) and \(X^{\z}_1 = x'\) equals \(\P(x \mid \z)\P(x' \mid \z)\). However, this probability can be more complex and depend on sophisticated factors beyond joint probability distributions, as restrictions or facilitations may make such changes more or less available. 
Thus, we can define:
\[
\PIEV_d^{\z}(X_{(x,x')} \to Y) := |g_{\text{in}}(x, \z) - g_{\text{in}}(x', \z)| \P(X^{\z}_0 = x, X^{\z}_1 = x')^d.
\]
\item 
More general than the previous generalization, when we incorporate the degree \(d\) as the exponent of the probability values, we are exponentially increasing the importance of more frequent treatment values. One could further adjust this importance by using other functions instead of exponential functions. Thus, we can generalize the definition as follows:
\[
\PIEV_d^{\z}(X_{(x,x')} \to Y) := |g_{\text{in}}(x, \z) - g_{\text{in}}(x', \z)| w_d(x_0, x_1),
\]
where \(w(x_0, x_1, d)\) is a weight function that allows for more manual control over the importance of treatment value changes (see \ref{binary DCE} ).
\item 
We can take the idea of previous item further and work with custom degrees. To do so, we assume a function \(D: \Supp(X) \times \Supp(X) \to \mathbb{R}\), and define:
\[
\PIEV_D^{\z}(X_{(x,x')} \to Y) := |g_{\text{in}}(x, \z) - g_{\text{in}}(x', \z)| w_{D(x_0, x_1)}(x_0, x_1 ),
\]
where \(D(x_0, x_1)\) is a custom degree function that controls the weight based on the values of \(x_0\) and \(x_1\).
The same idea can be applied to the degree \(r\) defined earlier. Indeed,  we  consider  \(r': \Supp(\Z) \to \mathbb{R}\) and define the following ($\Z\sim Q$):
\[
Q(\z) := \frac{\P(\z)^{r'(\z)}}{\sum_{\z' \in \Supp(\Z)} \P(\z')^{r'(\z')}}.
\]
\end{itemize}

\section{Limitations and Potential Drawbacks}
A main controversial  consequence of our methodology is that the PACE of $X$ on $Y$  depends on the variables $\Z$ that we are keeping constant. Hence, if we replace these variables $\Z$ with their parents $\W$ and update our causal relationships, then PACE would change. This is because the "natural availability of changing  $X$ values keeping $\Z$ constant" is replaced with the "natural availability of changing  $X$ values given $\W$ constant" (however, one may think of this as a property and not a drawback!)

Another drawback is the challenge in defining the indirect causal effect while ensuring that the total causal effect, as defined Equation in \ref{totaleffect}, can be expressed as the sum of the direct and indirect causal effects. This challenge arises primarily from the variability in the natural availability of changing the treatment values when we adjust the covariate that we control for, as highlighted in the first drawback mentioned above.

Furthermore, the previously mentioned random vector \(\Z\) may contain components that are difficult to intervene on, as we do in our direct causal effect metrics. In such cases, one could consider conditioning on these variables instead of intervening on them. This introduces some other versions of our causal metrics. In any case, the interpretation of each metric should align with the manner in which it was defined.

\section{Conclusion}

In this paper, we introduced a new causal inference point of view to investigate causality via direct causal effects. Our methodology includes a direct causal effect metric called PACE, which  is an integration of two concepts: 1) intervention (via total variation), and 2) natural availability of changing  the exposure/treatment values. The latter makes our methodology distinct from the other well-known frameworks for causal reasoning. Further, our causal methodology considers the rarity and frequency of events in observational studies, emphasizing their relevance to the core problem using the parameter $d$ in PACE and its variations such as PEACE, SPACE, and APACE. Furthermore, positive and negative versions of our causal metrics were defined to capture respective changes in outcomes, and a normalized version of PACE, called MEAN PACE, was provided. Similarly, the mean versions of other causal metrics introduced in the paper were also defined. In general, if for a given  covariate value, the MEAN PACE is dominated by terms related to more frequent (hypothetical) transitions between treatment levels, then MEAN PACE for that covariate value increases as the degree $d$ rises. Otherwise, it decreases. 
The paper also introduces an identifiability criterion for our causal metrics to address counterfactuals and provides generalizations of the approach from different perspectives. Finally, we compare our methodology with existing causal frameworks through various examples. Notably, our methodology has been successfully applied in two distinct areas: 1) Medical Sciences (see \cite{saki2024differentiating}), and 2) Reinforcement Learning (see \cite{khelifi2024causal}).





\bibliographystyle{elsarticle-num} 
\bibliography{bibliography}


\appendix

\section{Janzing et al. Causal Framework}\label{janz}
In this framework, five primary assumptions for a DAG 
$H$ are considered
as follows: 1) the local Markov assumption remains satisfied when one removes a set 
$\mathcal{S}$ 
of arrows in 
$H$ 
whose causal strength is 
0, 
2) if 
$H$
consists of exactly one arrow 
$X\to Y$, then the causal strength of this arrow equals 
$\I(X;Y)$, 3) the causal strength of 
$X\to Y$
only depends on 
$\P(Y|PA_Y)$
and 
$\P(PA_Y)$, where $PA_Y$ denotes the parents of $Y$,  4) the causal strength of 
$X\to Y$
is at least $\I(X;Y|PA^X_Y)$, where $PA^X_Y$
denotes all parents of $Y$
other than 
$X$,
and 5) if the causal strength of a set 
$\mathcal{S}$ of arrows is 
0, and 
$\mathcal{T}\subseteq \mathcal{S}$, 
then the causal strength of 
$\mathcal{T}$
is 
0  as well. 
According to the above five assumptions, to define the causal strength of an arrow 
$X\to Y$
in 
$H$, 
Janzing et al. considered the nodes of 
$H$
as electronic devices and the arrows as wires connecting the devices. The causal strength of $X\to Y$ is interpreted as the effect of cutting the wire. Next, a post-cutting DAG 
$H_{X\to Y}$
is obtained in which 
$Y$
is probabilistically fed with the values of 
$X$ 
independent of all other variables. That is if $X_1,\ldots,X_n$ are all nodes of $H$ (including $X$ and $Y$), then
\[\P_{X\to Y}(x_1,\ldots,x_n)=\prod_{i=1}^n\P_{X\to Y}(x_i | pa^x_{x_i}), \quad \P_{X\to Y}(x_i|pa^x_{x_i})=\sum_{X=x'}\P(x_i|pa^x_{x_i},x')P(x').\]
Similarly, for a set 
$\mathcal{S}$ 
of arrows in 
$H$,
one could define  
\[\P_{\mathcal{S}}(x_1,\ldots,x_n)=\prod_{i=1}^n\P_{\mathcal{S}}(x_i | pa^{\mathcal{S}}_{x_i}),\quad \P_{\mathcal{S}}(x_i|pa^{\mathcal{S}}_{x_i})=\sum_{\alpha}\P(x_i|pa^{\mathcal{S}}_{x_i},\alpha)\widetilde{\P}(\alpha),\]
and 
$\widetilde{\P}$
is the independent joint probability distribution of all source variables of arrows in 
$\mathcal{S}$,
and 
$\alpha$
varies among all values of the Cartesian product of these source variables. Finally, the authors defined the causal strength of 
$\mathcal{S}$
in 
$H$
to be 
$D_{\mathrm{KL}}(\P|| \P_{\mathcal{S}})$.

\section{Variational DCEs for Indirect Causes}\label{indirect}
Let us consider  $Y=g(X,\Z)$ and $X=h(\Z', \W)$. Now, we can use the composition of functions to calculate the variational DCE of $\W$ on $Y$ with respect to a new SEM. Indeed, the new SEM is the old one with two changes: 1) we remove $X=h(\Z',\W)$, and 2) we replace $Y=g(X,\Z)$ with $Y=\widetilde{g}(\Z,\W)=g(h(\Z',\W),\Z)$. An example is illustrated in Figure \ref{causalgraphalongpath}. In this example, since the value of $W$ is determined by the value of $Z$, we have that $\P(w|z)P(w'|z)=0$ for any $Z=z$ and $w,w'\in\Supp(W)$ with $w\neq w'$. However, this does not always happen when there is another random variable $E$ in such a way that $W$ is a function of $Z$ and $E$. 
\begin{figure}
\begin{center}
	\subfloat[\raggedright \scriptsize{The causal graph associated to the following SEM:
		$W=f(Z),
		X=h(W),
		Y=g(X,Z)
		$.}\label{causalgraphwithX}]{
		\pgfplotsset{compat=1.15}
		\begin{tikzpicture}[line cap=round,line join=round,>=triangle 45,x=1cm,y=1cm,scale =0.23]
			\clip(-13,-5.331741725374487) rectangle (14.827436373043655,10.284728569893227);
			\draw [line width=0.4pt] (6,0) circle (1.9915069670980314cm);
			\draw[blue] (6,0)node{$Y$};
			\draw [line width=0.4pt] (0,0) circle (1.9807036386815988cm);
			\draw[blue] (0,0)node{$X$};
			\draw [line width=0.4pt] (-6,0) circle (1.963810808206379cm);
			\draw[blue] (-6,0)node{$W$};
			\draw [line width=0.4pt] (0,6) circle (1.982772783282884cm);
			\draw[blue] (0,6)node{$Z$};
			\draw [->,line width=0.4pt] (-1.8808067583849193,5.372342571348855) --node[midway, above left, red]{$f$} (-5.313053703997356,1.8397439161025755);
			\draw [->,line width=0.4pt] (-4.036189191793621,0) --node[midway, below, red]{$h$} (-2,0);
			\draw [->,line width=0.4pt] (1.9807036386815988,0) --node[midway, below, red]{$g$} (4,0);
			\draw [->,line width=0.4pt] (1.8921793752273879,5.407509390714547) --node[midway, above right, red]{$g$} (5.272022006829138,1.8536849898132433);
	\end{tikzpicture}}
	\subfloat[\raggedright\scriptsize{The causal graph associated to the SEM, which is obtained by removing $X$ from the  SEM in the caption of Figure \ref{causalgraphwithX}:
		$W=f(Z), Y=\widetilde{g}(Z,W)=g(h(w), Z)$.}]{
		\pgfplotsset{compat=1.15}
		\begin{tikzpicture}[line cap=round,line join=round,>=triangle 45,x=1cm,y=1cm, scale = 0.23]
			\clip(-13.82,-5.331741725374487) rectangle (14.827436373043655,10.284728569893227);
			\draw [line width=0.4pt] (6,0) circle (1.9915069670980314cm);
			\draw[blue](6,0)node{$Y$};
			\draw [line width=0.4pt] (-6,0) circle (1.963810808206379cm);
			\draw[blue](-6,0)node{$W$};
			\draw [line width=0.4pt] (0,6) circle (1.982772783282884cm);
			\draw[blue](0,6)node{$Z$};
			\draw [->,line width=0.4pt] (-1.8808067583849193,5.372342571348855) --node[midway,above left,red]{$f$} (-5.313053703997356,1.8397439161025755);
			\draw [->,line width=0.4pt] (1.8921793752273879,5.407509390714547) --node[midway,above right,red]{$\widetilde{g}$} (5.272022006829138,1.8536849898132433);
			\draw [->,line width=0.4pt] (-4.036189191793621,0) -- node[midway,below,red]{$\widetilde{g}$} (4,0);
	\end{tikzpicture}}
	\caption{\raggedright \small The procedure to find the variational DCE of a variable along a path toward the outcome.}\label{causalgraphalongpath}
\end{center}
\end{figure}
\begin{Remark}
Assume that $\Z=H(\bm{\mathrm{V}})$, where $\bm{\mathrm{V}}$ might include $X$ or not. Then, we have that $Y=\widetilde{g}(X,\bm{\mathrm{V}})=g(X, H(\bm{\mathrm{V}}))$. Note that due to the possibility of having different natural availability of changing $\Z$ and $\bm{\mathrm{V}}$ values, each of the variational DCEs of $X$ on $Y$  might provide different values before and after replacing $\Z$ with $H(\bm{\mathrm{V}})$.
\end{Remark}

\section{A General Matrix Representation to Define Variational DCEs}\label{matrix}
In this appendix, we obtain matrix representations for our previously defined variational DCE metrics and show how this type of representation can help us define new variational DCE metrics. To do so, set
\[\alpha_i = \P(x_i|\z), \quad u_{ij}=|g_{in}(x_i,\z)-g_{in}(x_j,\z)|,\qquad 0\le i,j\le l.\]
We have that
\begin{align*}
\PIEV^{\z}(X\to Y) = \sum_{i=1}^lu_{i,i-1}\alpha_i\alpha_{i-1}=\sum_{i,j=0}^lu_{ij}\alpha_i\alpha_j\delta_{i,j+1}=v_X^t A v_X,
\end{align*}
where
\[
\delta_{l,k}=\left\{\begin{array}{ll}
1,& l=k\\0,& l\neq k
\end{array}\right.,\qquad v_X^t = \left[\begin{array}{ccc}\alpha_0&\cdots&\alpha_l\end{array}\right],\qquad A=(A_{ij}),\;\; A_{ij}=\delta_{i,j+1}u_{ij}.\]
Therefore, we have that
\[\PIEV^{\z}(X\to Y)=\textbf{[}\begin{array}{ccccc}\alpha_0&\alpha_1&\cdots&\alpha_{l-1}&\alpha_l\end{array}\textbf{]}\left[\begin{array}{ccccc} 0 & u_{01} & 0 & \cdots & 0\\
0 & 0 & u_{12} & \cdots & 0\\ 
\vdots & \vdots & \vdots & \ddots & \vdots\\
0 & 0 & 0 & \cdots & u_{l-1,l}\\
0 & 0 & 0 & \cdots & 0
\end{array}\right]\left[\begin{array}{c} \alpha_0\\\alpha_1\\\vdots\\\alpha_{l-1}\\\alpha_l \end{array}\right]\]
Now, one could see that 
\[\PIEV^{\z}_d(X\to Y)=\left(v_X^d\right)^t A v_X^d,\qquad \left(v_X^d\right)^t = \left[\begin{array}{ccc}\alpha_0^d&\cdots&\alpha_l^d\end{array}\right]. \]
In general, let $P\in\mathcal{P}(X)$. Then, we have that
\[\PIEV_d^{\z}(X_P\to Y)=\sum_{i,j=0}^l u_{ij}\alpha_i^d\alpha_j^d\delta_{ij}^{(P)}=\left(v_X^d\right)^tA^{(P)}v_X^d,\]
where 
\[\delta_{ij}^{(P)}=\left\{\begin{array}{lc} 1, & \exists\; 0\le k \le n_P\;\; x_i = x_{k+1}^{(P)},\, x_{j}=x_k^{(P)}\\ 0,& \text{otherwise} \end{array}\right.,\quad A^{(P)}=(A^{(P)}_{ij}),\;\;A^{(P)}_{ij}= \delta^{(P)}_{ij}u_{ij}. \]
In particular, if $P=(x_i,x_j)$ with $0\le i<j\le l$, then $A^{(P)}=u_{ij}e_{ij}$, where $e_{ij}$ is an $(l+1)\times (l+1)$ matrix whose all entries are 0 except for its $ij^{\mathrm{th}}$ entry which is $1$. Therefore, we have that
\begin{align*}
\PIV_d^{\z}(X\to Y) &= \max_{P\in\mathcal{P}(X)}\left(v_X^d\right)^tA^{(P)}v_X^d,\\
\SPIV_d^{\z}(X\to Y) &= \max_{\substack{x,x'\in\Supp(X)\\x<x'}}\left(v_X^d\right)^tA^{(x,x')}v_X^d.
\end{align*}
Moreover, we have that 
\[\APIV_d^{\z}(X\to Y) = \sum_{\substack{i,j=1\\i<j}}u_{ij}\alpha_i^d\alpha_j^d= \left(v_X^d\right)^tA^{all}v_X^d,\]
where $A^{all}=\left( A^{all}_{ij}\right)$ with $A^{all}_{ij}=u_{ij}$ when $i<j$ and $0$ otherwise.

Now, let $\Gamma_l$ be the set of all  $(l+1)\times(l+1)$ upper triangular matrices whose  $ij^{\text{th}}$ entry is $u_{ij}$ or 0 for any $0\le i<j<l$, and their diagonals are 0.  Assume that $\Lambda_l$ is a non-empty subset of $\Gamma_l$. Also, for any $B\in\Lambda_l$ and $P\in\mathcal{P}(X)$, let $B^{(P)}\in\Gamma_l$ correspond to $X_P$. Then, for any $P\in\mathcal{P}(X)$,  we define the probabilistic interventional $\Lambda_l$-variation of degree $d$ of $Y$ with respect to $X_P$ given $\Z=\z$ as follows:
\[\Lambda_l-\PIV_d^{\z}(X_P\to Y):=\max_{B\in\Lambda_l}\left(v_X^d\right)^tB^{(P)}v_X^d.\]
Here, for any $P\in\mathcal{P}(X)$ with $1\le k+1\le l+1$ elements, we need a compatibility property as follows:
\[\Lambda_l-\PIV_d^{\z}(X_P\to Y) = \Lambda_k-\PIV_d^{\z}(X_P\to Y),\]
while 
\[\Lambda_k-\PIV_d^{\z}(X_P\to Y)=\max_{B\in\Lambda_k}\left(w_X^d\right)^tBw_X^d,\quad w_X^d=\left[\begin{array}{ccc} \P(x_0^{(P)}|\z)^d &\cdots& \P(x_{k}^{(P)}|\z)^d\end{array}\right].\]

\section{Proofs of Results}\label{Proofs of Results}
\subsection{Theorem 3.2}\label{3.2}
\textbf{P1}: The probability distribution of $\Z$ could be obtained from the joint probability distribution of $X$ and $\bm{Z}$, and hence it is enough to show that $\PIV_d^{\z}(X\to Y)$ satisfies \textbf{P1}. Thus, it is enough to show that $\PIEV_d^{\z}(X_P\to Y)$ satisfies \textbf{P1} for any $P\in\mathcal{P}(X)$. The $i^{\mathrm{th}}$ term of $\PIEV_d^{\z}(X_P\to Y)$ is as follows:
\[A_i =\left|g_{in}(x_{i}^{(P)},\z)-g_{in}(x_{i-1}^{(P)},\z)\right|\P(x_{i}^{(P)}|\z)^d\P(x_{i-1}^{(P)}|\z)^d. \]
Clearly, $\P(x_{i}^{(P)}|\z)^d\P(x_{i-1}^{(P)}|\z)^d$ is determined by $d$ and the joint probability distribution of $X$ and $\Z$.  It remains $\left|g_{in}(x_{i}^{(P)},\z)-g_{in}(x_{i-1}^{(P)},\z)\right|$, which depends only on $g_{in}$. Therefore, $\PACE_d(X\to Y)$ satisfies Postulate \textbf{P1}. 

\vspace*{0.2cm}
\noindent \textbf{P2}: Let $P\in\mathcal{P}(X)$. It is enough to show that $\PIV_d^{\z}(X_P\to Y)\le\PIV_d^{\z}(X\to Y)$. To do so, we note that $\mathcal{P}(X_P)\subseteq\mathcal{P}(X)$, and hence
\begin{align*}
\PIV_d^{\z}(X_P\to Y)&=\max\left\{\mathcal{P}\mathcal{I}\mathcal{E}\mathcal{V}^{\bm{z}}((X_P)_Q\to Y): Q\in\mathcal{P}(X_P)\right\}\\
&=\max\left\{\mathcal{P}\mathcal{I}\mathcal{E}\mathcal{V}^{\bm{z}}(X_Q\to Y): Q\in\mathcal{P}(X_P)\right\}\\
&\le \max\left\{\mathcal{P}\mathcal{I}\mathcal{E}\mathcal{V}^{\bm{z}}(X_Q\to Y): Q\in\mathcal{P}(X)\right\}=\PIV_d^{\z}(X\to Y).
\end{align*}

\vspace*{0.2cm}
\noindent \textbf{P3}: In fact, the controlled PACE of $X$ on $Y$ given $\Z=\z$ is $\PIV_d^{\z}(X\to Y)$, and  we have that 
\[	\PACE_d(X\to Y)=\E_{\Z}\left(\mathcal{P}\mathcal{I}\mathcal{V}_d^{\z}(X\to Y)\right).\]
Thus, Postulate \textbf{P3} is satisfied.

\vspace*{0.2cm}
\noindent \textbf{P4}: First, assume that $d>0$ and  $\Z=\z$ are given, and  we do not have any interventional changes of $Y$ value with respect to changing $X$ values while keeping $\Z=\z$ constant. We show that $\PIV_d^{\z}(X\to Y)=0$. To do so, let $P\in \mathcal{P}(X)$. Then,   since $\left|g_{in}(x_{i}^{(P)},\z)-g_{in}(x_{i-1}^{(P)},\z)\right|=0$, the $i^{\mathrm{th}}$ term of $\PIEV_d^{\z}(X_P\to Y)$ is 0 for any $1\le i\le n_P$. It follows that $\PIEV_d^{\z}(X_P\to Y)=0$, and hence $\PIV_d^{\z}(X\to Y)=0$. On the other hand, assume that $\PIV_d^{\z}(X\to Y)=0$. Let $1\le i<j\le l$ be arbitrary. Then, we have that 
\[|g_{in}(x_j,\z)-g_{in}(x_i,\z)|\P(x_j|\z)^d\P(x_i|\z)^d=\PIEV_d^{\z}(X_{P_0}\to Y)\le \PIV_d^{\z}(X\to Y)=0,\]
where $P_0=(x_i,x_j)\in\mathcal{P}(X)$. Thus, when the  change from $X=x_i$ to $X=x_j$ (or vice versa) is possible, then $\P(x_j|\z)^d\P(x_i|\z)^d\neq 0$, which implies that $|g_{in}(x_j,\z)-g_{in}(x_i,\z)|=0$. Therefore, the value of $Y$ remains constant.

\vspace*{0.2cm}
\noindent \textbf{P5}: it is a direct result of the definition of $\PACE_d(X\to Y)$.
\subsection{ Proposition 3.4}\label{3.4}
On the contrary, assume that $\PACE_d(X\to Y)=0$ but $\PACE_d(W\to Y)\neq 0$. It follows that there exists $\z\in\Supp(\Z)$ with $\PIV_d^{\z}(W\to Y)\neq 0$, which implies that there exists $w_1,w_2\in\Supp(W)$ with $w_1<w_2$ in such  a way that $(w_1,\z)$ and  $(w_2,\z)$ are in the domain of $h$ and   $|\widetilde{g}_{in}(w_2,\z)-\widetilde{g}_{in}(w_1,\z)|\P(w_1|\z)^d\P(w_2|\z)^d\neq 0$. Now, let $x_i=h(w_i,\z)$ for $i=1,2$. Then, $\widetilde{g}_{in}(w_i,\z)=\widetilde{g}^{W,\Z}(w_i,\z)=g^{W,\Z}(x_i,\z)=g^{X, \Z}(x_i,\z)$ for $i=1,2$. It follows from $\P(w_1|\z)^d\P(w_2|\z)^d\neq 0$ that $\P(x_1|\z)^d\P(x_2|\z)^d\neq 0$, and hence $g^{X,\Z}(x_1,\z)=g^{X,\Z}(x_2,\z)$ by $\PACE_d(X\to Y)=0$, a contradiction!

\subsection{Theorem  3.10}\label{3.10}
It is enough to show the inequalities for their corresponding variation metrics.  We have that 
\[\PIEV_d^{\bm{z}}(X\to Y)=\PIEV_d^{\z}(X_{\Supp(X)}\to Y)\le\max_{P\in\mathcal{P}(X)}\PIEV_d^{\bm{z}}(X_P\to Y)=\PIV_d^{\bm{z}}(X\to Y).\]
Now, let $P\in\mathcal{P}(X)$. Then, obviously we have that 
\begin{align*}
\PIEV_d^{\bm{z}}(X_P\to Y)&=\sum_{i=1}^{n_P}|g_{in}(x_i^{(P)})-g_{in}(x_{i-1}^{(P)})|\P(x_i^{(P)}|\bm{z})^d\P(x_{i-1}^{(P)}|\bm{z})^d\\
&\le \sum_{\substack{x,x'\in\Supp(X)\\ x< x'}}|g_{in}(x)-g_{in}(x')|\P(x|\bm{z})^d\P(x'|\bm{z})^d= \APIV_d^{\bm{z}}(X\to Y).
\end{align*}
It follows that 
\[\PIV_d^{\z}(X\to Y) = \max_{P\in\mathcal{P}(X)}\PIEV_d^{\bm{z}}(X_P\to Y)\le \APIV_d^{\bm{z}}(X\to Y). \]
Thus, we proved the first item. To prove the second item, let $x,x'\in\Supp(X)$ with $x<x'$. Then, 
\begin{align*}
\PIEV_d^{\z}(X_{(x,x')}\to Y)\le \max_{P\in\mathcal{P}(X)}\PIEV_d^{\z}(X_P\to Y)=\PIV_d^{\z}(X\to Y),
\end{align*}
which implies that
\begin{align*}
\SPIV_d^{\bm{z}}(X\to Y)&=\max_{\substack{\alpha,\beta\in\Supp(X)\\\alpha<\beta}} \PIEV_d^{\z}(X_{(\alpha,\beta)}\to Y)\le \PIV_d^{\z}(X\to Y).
\end{align*}
The remained inequality in the second item is included in the first item's inequalities that were shown above.
\subsection{Proposition 3.11}\label{infty}
Let $\epsilon=\min\left\{|g_{in}(\alpha,\bm{z})- g_{in}(\beta,\bm{z})|/2, f(\beta|\bm{z})^d/2\right\}$. Then, it follows from the continuity of $g_{in}(\cdot,\bm{z})$ and $f^d(\cdot|\bm{z})$ that there exists $\delta>0$ such that for any $x$ with $|x-\beta|<\delta$, we have that $|f(x|\bm{z})^d-f(\beta|\bm{z})^d|<\epsilon$ and $|g_{in}(x,\bm{z})-g_{in}(\beta,\bm{z})|<\epsilon$. Now, without loss of generality assume that $\alpha<\beta$. For any positive integer $n$, define $A_n\subseteq\Supp(X)$ as 
$A_n=\{\alpha, x_1,\ldots, x_n\}$, 
where  $|x_i-\beta|<\delta$ for any $1\le i\le n$. Now, by considering $L=\APIV_d^{\bm{z}}(X_{A_n}\to Y)$, we have that
\begin{align*}
L\ge& \sum_{i=1}^n|g_{in}(x_i,\bm{z})-g_{in}(\alpha,\bm{z})|f(x_i|\bm{z})^df(\alpha|\bm{z})^d\\
\ge & \sum_{i=1}^n(|g_{in}(\alpha,\bm{z})-g_{in}(\beta,\bm{z})|-|g_{in}(x_i,\bm{z})-g_{in}(\beta,\bm{z})|)(f(\beta|\bm{z})^d-\epsilon)f(\alpha|\bm{z})^d\\
>&\sum_{i=1}^n\left(|g_{in}(\alpha,\bm{z})-g_{in}(\beta,\bm{z})|-\frac{|g_{in}(\alpha,\bm{z})-g_{in}(\beta,\bm{z})|}{2}\right)\left(f(\beta|\bm{z})^d-\frac{f(\beta|\bm{z})^d}{2}\right)f(\alpha|\bm{z})^d\\
=&n\left(\frac{(|g_{in}(\alpha,\bm{z})-g_{in}(\beta,\bm{z})|)f(\alpha|\bm{z})^{d}f(\beta|\bm{z})^{d}}{4}\right).
\end{align*}
Therefore, we have that 
\begin{align*}
\APIV_d^{\bm{z}}(X\to Y)&\ge \lim_{n\to \infty}\APIV_d^{\bm{z}}(X_{A_n}\to Y)\\
&\ge \lim_{n\to\infty}n\left(\frac{(|g_{in}(\alpha,\bm{z})-g_{in}(\beta,\bm{z})|)f(\alpha|\bm{z})^{d}f(\beta|\bm{z})^{d}}{4}\right)=\infty,
\end{align*}
which implies that 
$\APIV_d^{\bm{z}}(X\to Y)=\infty$.

\subsection{Theorem 3.18}\label{3.18}
First, we show that $\PIEV_d^{(\z,\u_Y)}(X_{(x,x')}\to Y)$ is identifiable. By Assumption~(1), we have
\begin{align*}
g_{in}(x,\z,\u_Y)-g_{in}(x',\z,\u_Y)&=\left(g_{in}^{(1)}(x,\z)+g_{in}^{(2)}(\z,\u_Y)\right)-\left(g_{in}^{(1)}(x',\z)+g_{in}^{(2)}(\z,\u_Y)\right)\\
&=g_{in}^{(1)}(x,\z) - g_{in}^{(1)}(x',\z),
\end{align*}
which is identifiable. To be done with the first step, it is enough to show that $\P(x|\z,\u_Y)$ is identifiable for any $\left(X,\Z,\U_Y\right)=(x,\z,\u_Y)$. By Assumption (2), $\P(x|\z,\u_Y)=\P(x|\z)$, which means that $\P(x|\z,\u_Y)$ is identifiable. Thus, $\PIEV_d^{(\z,\u_Y)}(X_{(x,x')}\to Y)$ is identifiable. Since by the above identifiability, $\PIEV_d^{(\z,\u_Y)}(X_{(x,x')}\to Y)$ does not depend on $\U_Y$, we have that
\begin{align*}
\PEACE_d(X_{(x,x')}\to Y)&=\E_{(\Z,\U_Y)}\left(\PIEV_d^{(\z,\u_Y)}(X_{(x,x')}\to Y)\right)\\
&= \E_{\Z}\left(\PIEV_d^{(\z,\u_Y)}(X_{(x,x')}\to Y)\right),
\end{align*}
and hence $\PEACE_d(X_{(x,x')}\to Y)$ is identifiable. 
It follows from Assumption (1) that
\begin{small}\begin{align*}
	\E(Y_{x',\z}-Y_{x,\z})&=\E\left(g_{in}^{(1)}(x',\z)- g_{in}^{(1)}(x,\z)\right)+\E\left(g_{in}^{(2)}\left(\z,\left(U_Y\right)_{x',\z}\right)-g_{in}^{(2)}\left(\z,\left(U_Y\right)_{x,\z}\right)\right).
	\end{align*}\end{small}
	Hence, by the other assumptions, $\E(Y_{x',\z}-Y_{x,\z})- \left(g_{in}^{(1)}(x',\z)- g_{in}^{(1)}(x,\z)\right)$ equals
	\begin{align*}
&\E\left(g_{in}^{(2)}\left(\z,\left(U_Y\right)_{x',\z}\right)\right)-\E\left(g_{in}^{(2)}\left(\z,\left(U_Y\right)_{x,\z}\right)\right)\\
=& \E\left(g_{in}^{(2)}\left(\z,\left(U_Y\right)_{x',\z}\right)\middle|\Z=\z\right)-\E\left(g_{in}^{(2)}\left(\z,\left(U_Y\right)_{x,\z}\right)\middle| \Z=\z\right)\\
=&\E\left(g_{in}^{(2)}\left(\z,\left(U_Y\right)_{x',\z}\right)\middle|X=x',\Z=\z\right)-\E\left(g_{in}^{(2)}\left(\z,\left(U_Y\right)_{x,\z}\right)\middle| X=x,\Z=\z\right)\\
=&\E\left(g_{in}^{(2)}\left(\z,U_Y\right)\middle|X=x',\Z=\z\right)-\E\left(g_{in}^{(2)}\left(\z,U_Y\right)\middle|X=x,\Z=\z\right)\\
=&\E\left(g_{in}^{(2)}\left(\z,U_Y\right)\middle|\Z=\z\right)-\E\left(g_{in}^{(2)}\left(\z,U_Y\right)\middle|\Z=\z\right)=0.
\end{align*}
Thus, we have that
$\E(Y_{x',\z}-Y_{x,\z})=g_{in}^{(1)}(x',\z)- g_{in}^{(1)}(x,\z)$. 
Now, it follows from Assumption~(3) and Assumption (5) that
$\E(Y_{x,\z})=\E(Y_{x,\z}|\Z=\z)=\E(Y_{x,\z}|X=x, \Z=\z)$. 
Moreover, $\E(Y_{x,\z}|X=x, \Z=\z)=\E(Y|X=x,\Z=\z)$, which implies that
$\E(Y_{x,\z})=\E(Y|X=x,\Z=\z)$. 
Finally, we have that 
\small{\[\PIEV_d^{(\z,\u_Y)}(X_{(x,x')}\to Y) = \bigl|\E(Y|X=x',\Z=\z)-\E(Y|X=x,\Z=\z)\bigr|\P(x'|\z)^d\P(x|\z)^d,\]}
and hence, the proof is completed.

\section{Examples}\label{different}
\subsection{Counterexample for Remark 3.6}
Let $\mathrm{Supp}(X)=\{0,1,2\}$ and \[Y=g(X)=\left\{\begin{array}{ll} 0,& X=0\\ 2,&X=1\\ 1,& X=2\end{array}\right.,\qquad \P_X(0)=\P_X(2)=\frac{16}{35},\quad\P_X(1)=\frac{3}{35}.\]
Assume that $P=\{0,1,2\}$ and $Q=\{0,2\}$. Then, we have that 
\begin{align*}
\PIV_d(X_P\to Y)&=3\left(\frac{48}{35^2}\right)^d,\quad \PIV_d(X_Q\to Y)=
\left(\frac{16}{35}\right)^{2d}.
\end{align*}
The above variations are plotted in Figure \ref{figureofexampledifferentddifferentpartitions}. 

\pgfplotsset{compat = newest}
\begin{figure}
\begin{center}
	\caption{\raggedright \small{The red and the blue graphs show $\PIV_d(X_P\to Y)$ and $\PIV_d(X_Q\to Y)$, where $P=\{0,1,2\}$ and $Q=\{0,2\}$.}}\label{figureofexampledifferentddifferentpartitions}
	\begin{tikzpicture}
		\begin{axis}[
			xmin = 0, xmax = 2,
			ymin = 0, ymax = 3.5,
			scale=0.7,
			xlabel = $d$, ylabel = $\PIV_d(X_{P'}\to Y)$]
			\addplot[
			domain = 0:2,   samples = 200,
			smooth,
			thick,
			blue,
			] {3*(48/35^2)^x};
			\addplot[
			domain = 0:2,   samples = 200,
			smooth,
			thick,
			red,
			] {(16/35)^(2*x)};
			
			\legend{
				$P' = P$,
				$P' = Q$
			}
		\end{axis}
	\end{tikzpicture}
\end{center}
\end{figure}
\subsection{Example of Strict Inequalities for Theorem 3.10 }\label{ex3.10}
Let $X$ be a random variable with $\Supp(X)=\{1,2,3,4\}$ in such a way that
\[\P_X(1)=\frac{1}{6},\quad \P_X(2)=\frac{1}{12},\quad \P_X(3)=\frac{1}{4},\quad \P_X(4)=\frac{1}{2}.\]	Also, assume that 	\[Y=g(X)=\left\{\begin{array}{ll} X,& X\in\{1,2,3\}\\ 1,& X=4\end{array}\right.\]
Then, one could see that
\begin{align*}
\PACE(X\to Y)&=\frac{1}{3},\quad \PEACE(X\to Y)=\frac{41}{144}, 
\\	
\APACE(X\to Y)&=\frac{59}{144},\quad \SPACE(X\to Y)=\frac{1}{4}.
\end{align*}
In the above example, one could see that also for different degrees $d>0$, we have that
\begin{align*}
\PEACE_d(X\to Y)&<\PACE_d(X\to Y)<\APACE_{d}(X\to Y),\\
\SPACE_d(X\to Y)& <\PEACE_d(X\to Y).
\end{align*}

\subsection{Example of Strict Inequalities for Corollary 3.14}\label{ex3.14}
Let $X$ be a random variable with $\Supp(X)=\{1,2,3\}$ and $d>0$ be given. Assume that  
\[\P_X(1)=\P_X(3)=\frac{4^{\frac{1}{d}}}{2\times 4^{\frac{1}{d}}+1},\quad \P_X(2)=\frac{1}{2\times 4^{\frac{1}{d}}+1}.\]
Also, assume that 
\[Y=g(X)=\left\{\begin{array}{ll}2X,& X\in\{1,2\}\\ 1,&X=3\end{array}\right..\]
One could see that
\begin{align*}
\PIV_d(X\to Y)&=\frac{20}{\left(2\times 4^{\frac{1}{d}}+1\right)^{2d}},\quad \PIV_d(X\to Y)^{\bm{+}}=\frac{8}{\left(2\times 4^{\frac{1}{d}}+1\right)^{2d}}, \\	
\PIV_d(X\to Y)^{\bm{-}}&=\frac{16}{\left(2\times 4^{\frac{1}{d}}+1\right)^{2d}}.
\end{align*}
Clearly, in this example, we have that
\begin{align*}
&\PIV_d^{\bm{z}}(X\to Y)< \PIV_d^{\bm{z}}(X\to  Y)^{\bm{+}}+\PIV_d^{\bm{z}}(X\to Y)^{\bm{-}},\\
&\PIV_d^{\bm{z}}(X\to Y)>\max\{ \PIV_d^{\bm{z}}(X\to  Y)^{\bm{+}},\PIV_d^{\bm{z}}(X\to Y)^{\bm{-}}\}.
\end{align*}
It follows that 
\begin{align*}
&\PACE_d(X\to Y)<\PACE_d(X\to Y)^{\bm{+}}+\PACE_d(X\to Y)^{\bm{-}},\\
&\PACE_d(X\to Y)>\max\{\PACE_d(X\to Y)^{\bm{+}},\PACE_d(X\to Y)^{\bm{-}}\}.
\end{align*}
\end{document}